# Graph Signal Processing for Heterogeneous Change Detection—Part I: Vertex Domain Filtering

Yuli Sun, Lin Lei, Dongdong Guan, Gangyao Kuang, *Senior Member, IEEE*, Li Liu, *Senior Member, IEEE*

***Abstract*—This paper provides a new strategy for the Heterogeneous Change Detection (HCD) problem: solving HCD from the perspective of Graph Signal Processing (GSP). We construct a graph for each image to capture the structure information, and treat each image as the graph signal. In this way, we convert the HCD into a GSP problem: a comparison of the responses of the two signals on different systems defined on the two graphs, which attempts to find structural differences (Part I) and signal differences (Part II) due to the changes between heterogeneous images. In this first part, we analyze the HCD with GSP from the vertex domain. We first show that for the unchanged images, their structures are consistent, and then the outputs of the same signal on systems defined on the two graphs are similar. However, once a region has changed, the local structure of the image changes, *i.e.*, the connectivity of the vertex containing this region changes. Then, we can compare the output signals of the same input graph signal passing through filters defined on the two graphs to detect changes. We design different filters from the vertex domain, which can flexibly explore the high-order neighborhood information hidden in original graphs. We also analyze the detrimental effects of changing regions on the change detection results from the viewpoint of signal propagation. Experiments conducted on seven real data sets show the effectiveness of the vertex domain filtering based HCD method.**

***Index Terms*—Heterogeneous change detection, graph signal processing, vertex domain, graph, multimodal.**

## I. Introduction

As a very important task in remote sensing, Change Detection (CD) aims at identifying the changes on Earth's surface by comparing multitemporal remote sensing images acquired over the same geographical area but at different times [1]. It has a wide range of real world applications, including environmental monitoring, land management, urban development and damage assessment [2]–[4]. Traditionally, most research in CD has been devoted to homogeneous CD which performs change detection with images acquired from the same sensor [5]–[7]. With the rapid development of high resolution imaging technique, increasingly more image data (such as Synthetic Aperture Radar (SAR) images and optical images) from different types of sensors can be collected conveniently, creating even more research opportunities for detecting and monitoring subtle changes of the Earth surface at a finer scale.

Y. Sun, L. Lei and G. Kuang are with College of Electronic science, National University of Defense Technology, Changsha 410073, China.

D. Guan is with the HighTech Institute of Xi'an, Xi'an 710025, China.

L. Liu is with the College of System Engineering, National University of Defense Technology, Changsha, 410073, China.

This work was supported in part by the National Key Research and Development Program of China No. 2021YFB3100800 and the National Natural Science Foundation of China under Grant Nos. 61872379, 12171481 and 61971426.

In recent years, Heterogeneous Change Detection (HCD) has emerged, starting to gain increasing attention. It addresses the problem of performing change detection with images coming from different sources like different types of sensors (such as a SAR image and an optical image), *i.e.*, multitemporal change detection with heterogeneous remote sensing images. There are several reasons account for this trend.

Firstly, as each of these sensors captures different aspects and characteristics of the Earth surface, HCD makes simultaneous use of multiple complementary data for detecting changes, overcoming the limitation of classical CD methods in relying on the availability of homogeneous data acquired by the same sensor, which may not be available sometimes due to the environmental conditions like bad light or weather conditions (*e.g.*, clouds, fogs and rain). Secondly, HCD can improve the temporal resolution when monitoring long term change evolution by inserting heterogeneous data [8]. Finally, HCD can shorten the response time of change analysis in case of sudden events (*e.g.*, floods and earthquakes) by using the first available images to assess the damages instead of waiting for the arrival of next homogeneous images [9].

In classical change detection, we can directly compare the given images (denoted as $x$ and $y$) to calculate the difference image (DI), such as via image differencing in the case of optical images [10], and image ratioing (or logarithmic ratioing) in the case of SAR images [11]. However, the multitemporal images in HCD are acquired by different sensors, and show quite different image characteristics [12]. The common assumption in classical CD that the multitemporal images measure the same quantities and show the similar statistical behavior is generally violated in HCD. As a result, most existing CD techniques do not directly apply for HCD. Therefore, the essential issue in HCD is how to make the "incomparable" images "comparable". To achieve this, a common solution is to tranform the heterogeneous images into a common domain $\mathcal{Z}$ as $\mathcal{M}_1 : x \to z$ and $\mathcal{M}_2 : y \to z'$.

### A. Related work

With the paradigm $DI = \mathcal{M}_1(x) \ominus \mathcal{M}_2(y)$, HCD methods are proposed with different types of $\mathcal{M}_1$ and $\mathcal{M}_2$, where $\ominus$ denotes the difference operator. Generally, we can classify algorithms differently around the conditions, techniques and domains of the transformation: according to whether labeled samples are required in the transformation process, HCD can be divided into supervised [13], [14], semisupervised [15], [16] and unsupervised [17]–[19]; according to the methods used for $\mathcal{M}_1$ and $\mathcal{M}_2$, HCD can be classified as traditional machine learning based [13], [20], [21] and deep learning based [22]–[24]; and according to the transformed common domain $\mathcal{Z}$, HCD can also be divided into classification comparison based, image regression based, and feature transformation based.

1) The classification comparison based methods first transform the images into a common category space by taking $\mathcal{M}_1$ and $\mathcal{M}_2$ as classifiers, and then compare the classification results to detect changes, such as the post-classification comparison method [14], the compound classification method [25], [26], and the classified adversarial network based method [27]. The advantages of such methods are that they are intuitive, robust to co-registration errors due to the object-wise comparison, and able to indicate the kind of change. However, classification comparison based methods are supervised or semi-supervised in order to train the accurate classifiers of $\mathcal{M}_1$ and $\mathcal{M}_2$, and they may suffer from the risk of accumulation of classification errors.

2) The image regression methods first transform one image (*e.g.*, $x$) to the domain of the other image (*e.g.*, $y$) by setting one transformation function ($\mathcal{M}_1$) to the identity matrix and the other ($\mathcal{M}_2$) to the image translation function, and thereby convert the issue into homogeneous CD. For example, some traditional methods construct the pixel-to-pixel mappings between heterogeneous images, such as the homogeneous pixel transformation method (HPT) [13], the affinity matrix based regression method [20]. Some deep translation methods have also been proposed, such as the image style transfer-based method (IST) [22], and the cycle-consistent generative adversarial networks (Cycle-GAN) based HCD methods [28]–[30].

3) The feature transformation methods transform the images into a common constructed or latent learned feature space by taking $\mathcal{M}_1$ and $\mathcal{M}_2$ as feature extraction operators. For example, some traditional methods manually construct the similarity measures that are assumed to be imaging-modality-invariant to calculate the changes, such as the copula theory based Kullback-Leibler (KL) distance [31], manifold learning method [21], kernel canonical correlation analysis (kCCA) method [32], nonlocal pixel pairwise-based method [33]. Some deep learning methods compare the images in the latent feature spaces that are learned by the deep neural networks, such as the symmetric convolutional coupling network [34], spatially self-paced convolutional network [35], self-supervised learning with pseudo-Siamese networks [18], [19], probabilistic model based on bipartite convolutional neural network [23], semi-supervised Siamese network [15], and the commonality autoencoder based method [24].

Although the above methods have achieved remarkable detection results in some HCD scenes, most of them still suffer from two main challenges.

- First, the connections between the heterogeneous images established by these methods are generally based on certain assumptions (*e.g.*, some imaging-modality-invariant assumptions in traditional methods) or trained transformations (*e.g.*, some networks in deep learning methods). These connections may be unstable and non-universal when the HCD scene is very complex (*e.g.*, diversity of ground objects, difference of imaging conditions), the noise in image is sever (especially the speckle noise in SAR image), or the training samples are not sufficient or mixed with wrong samples.
- Second, the negative influence of the unknown changed samples in the transformation is difficult to eliminate, both for training the transformation functions ($\mathcal{M}_1$ and $\mathcal{M}_2$) and completing the transformation process ($x \xrightarrow{\mathcal{M}_1} z$ and $y \xrightarrow{\mathcal{M}_2} z'$). In particular, this problem is rarely mentioned by other studies, partly for two reasons: 1) this challenge is unique to HCD, *i.e.*, it is not a problem in homogeneous CD that directly compares images without the design of transformations; 2) the previous methods usually treat HCD as a two-step process, transforming first and comparing latter, which tends to ignore the impact of the detected changes produced by the second process on the first transformation process. In order to alleviate this negative impact, an iterative framework that combines the two processes to perform a coarse-to-fine detection is needed [36].

### B. Motivations

The graph model can efficiently capture the structure information of an image, and the image processing on graphs has been proven to be effective by a large number of applications [37], [38], which has also been used in CD task. In the homogeneous CD of SAR images, a point-wise graph is constructed for the first image [39], and then the DI is calculated by superimposing two images on the same graph for comparison; a pixel-wise hypergraph is constructed for each image in [40], and then the DI is calculated by matching each vertex and hyperedge between the two hypergraphs. Here, the graphs (or hypergraphs) are used to incorporate the spatial-intensity information to resist the speckle noise in SAR images.

Recently, some graph based HCD methods have also been proposed. An approximate local graph is constructed by using Nystrom extension for each image [41], then the graphs are fused by minimizing the similarity between the graphs to detect the changes. Based on the self-similarity property, the patch-wise graphs [42] and the superpixel-wise graphs [43] are constructed to capture the structure of images, and then the graphs are compared to calculate the DI by graph projection or used to perform image regression. In addition, a fractal projection method has also been proposed for HCD based on the self-similarity property [44], which projects the pre-event image to the domain of post-even image with the fractal code of the pre-event image.

From the above analysis, we can find that the graph based methods have two attractive features: 1) these methods are intuitive, interpretable and very simple, which does not require a complex training process or any labelled samples; 2) the graph can capture the inherent structural information of an image, which is robust to the noise and shows the imaging-modality-invariant property that is ideal for HCD problem. Combining with the challenges of HCD analyzed above, two aspects of the graph based methods need to be considered in focus: first, how to adequately represent the image structure by graph; second, how to measure the structure difference in the presence of negative influence of unknown changed data. In addition, although the feasibility of these graph based methods has been experimentally verified, they have not so far been demonstrated theoretically in depth. Furthermore, there is no unified theoretical framework for these graph based HCD methods. These are exactly what this paper is devoted to exploring.

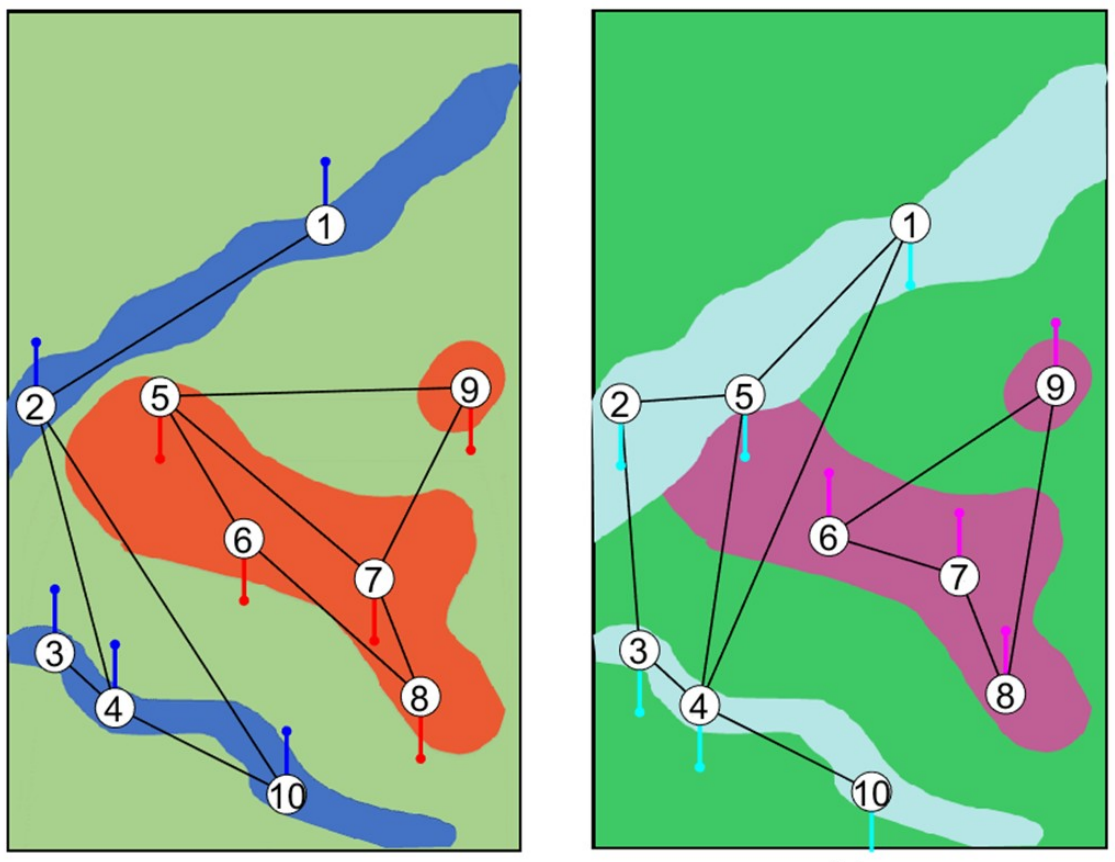

(a) (b)

Fig. 1. Graphs and graph signals for the pre-event image (a) and the post-event image (b). The structure information of the image is captured by the graph topology that represents the similarity relationships between objects. The image can be treated as the signal on the graph, and the graph signal values are represented by the vertical lines. Among the 10 vertices in the example, the area represented by vertex 5 has changed.

### C. Contributions

In this paper, we propose a new perspective for HCD, that is, converting the HCD problem as a graph signal processing (GSP) problem. We construct a graph for each image to capture the structure information, and then treat each image as the signal on the graph, as shown in Fig. 1. In this way, the changes between heterogeneous images caused by the event will manifest themselves in two aspects: the structure difference between graphs and the signal difference on the graph, as illustrated by Fig. 2. Thereby, we can compare the responses of the two signals (*i.e.*, images) on different systems (*i.e.*, filters) defined on the two graphs to detect the changes, from the vertex domain (part I) and the spectral domain (part II) [45] of GSP. The main contributions of this part are as follows.

- We convert the HCD into a GSP problem and propose a framework to solve the HCD by employing the principles of GSP on vertex domain.
- We analyze the negative influence of changes on the HCD from the perspective of signal propagation and use an iterative strategy to alleviate this influence.
- We design different filters for the HCD to explore the high-order neighborhood information hidden in original graphs, and give some discussions and extensions of the proposed method.
- Experimental results on seven real data sets demonstrate the effectiveness of the proposed method by comparing with some state-of-the-art (SOTA) methods (source code will be made available at https://github.com/yulisun/HCD-GSPpartI).

TABLE I
LIST OF IMPORTANT NOTATIONS.

| Symbol | Description |
|---|---|
| $\hat{\mathbf{X}}$, $\hat{\mathbf{Y}}$ | pre-event and post-even images |
| $\mathbf{X}$, $\mathbf{Y}$ | feature matrices of $\hat{\mathbf{X}}$, $\hat{\mathbf{Y}}$ |
| $\mathbf{X}_i$ | $i$-th **row** of matrix $\mathbf{X}$ |
| $x_{i,j}$ | $i$-th row and $j$-th column element of $\mathbf{X}$ |
| $G_{t1} = \{\mathcal{V}_{t1}, \mathcal{E}_{t1}, \mathbf{W}_{t1}\}$ | KNN Graph of the pre-event image |
| $G_{t2} = \{\mathcal{V}_{t2}, \mathcal{E}_{t2}, \mathbf{W}_{t2}\}$ | KNN Graph of the post-event image |
| $\mathbf{A}_{t1}, \mathbf{A}_{t2}$ | adjacent matrices |
| $\mathbf{D}_{t1}, \mathbf{D}_{t2}$ | degree matrices |
| $\mathbf{W}_{t1}, \mathbf{W}_{t2}$ | weight matrices |
| $\mathbf{P}_{t1}, \mathbf{P}_{t2}$ | random walk matrices |
| $\mathbf{L}_{t1}, \mathbf{L}_{t2}$ | Laplacian matrices |
| $\mathbf{I}_N$ | an $N \times N$ identity matrix |
| $\mathbf{1}_N$ | an $N \times 1$ column vector of ones |

### D. Outline

The remainder of this part is structured as follows: Section II describes the HCD from the perspective of GSP. Section III proposes a framework for HCD by using the vertex domain filtering. Section IV gives some discussions of the proposed method. Section V presents the experimental results. Finally, Section VI concludes this part and mentions the future work. For convenience, Table I lists some important notations used in the rest of this part. One notation should be noted: $\mathbf{X}_i$ represents the $i$-th **row** of matrix $\mathbf{X}$ in this paper, not the $i$-th column as commonly used.

## II. HCD From GSP Perspective

### A. Background of vertex filtering on graphs

Let $G = \{\mathcal{V}, \mathcal{E}, \mathbf{W}\}$ be a graph, and $\mathbf{f} = [f_1, \cdots, f_N]^T$ be the signal on the graph $G$, then a graph shift operator $\mathbf{S}$ is defined as *a local operation that replaces a signal value at each vertex with the linear combination of the signal values at the neighbors of that vertex* [46], [47]. Common choices for the graph shift operator are: 1) the adjacent matrix $\mathbf{A}$ or weight matrix $\mathbf{W}$; 2) the random walk (diffusion) matrix $\mathbf{P} = \mathbf{D}_w^{-1}\mathbf{W}$, where $\mathbf{D}_w$ represents the diagonal matrix with the $i$-th diagonal element being $\sum_{j=1}^{N} w_{i,j}$; 3) the Laplacian matrix $\mathbf{L} = \mathbf{D}_w - \mathbf{W}$, symmetric normalized Laplacian matrix $\mathbf{L}^{\text{sym}} = \mathbf{D}_w^{-1/2}\mathbf{L}\mathbf{D}_w^{-1/2}$, or random-walk normalized Laplacian matrix $\mathbf{L}^{\text{rw}} = \mathbf{D}_w^{-1}\mathbf{L}$.

A linear, shift-invariant system (filter) can be defined as the polynomials in the graph shift operator $\mathbf{S}$ of the form [48]–[50]

$$H(\mathbf{S}) = h_0\mathbf{S}^0 + h_1\mathbf{S} + \cdots + h_M\mathbf{S}^M = \sum_{m=0}^{M} h_m\mathbf{S}^m, \quad (1)$$

where $\mathbf{S}^0 = \mathbf{I}$, and $h_0, h_1, \cdots, h_M$ are system coefficients. The output signal of the system (1) with the input signal $\mathbf{f}_{\text{in}}$ is

$$\mathbf{f}_{\text{out}} = H(\mathbf{S})\,\mathbf{f}_{\text{in}} = \sum_{m=0}^{M} h_m \mathbf{S}^m \mathbf{f}_{\text{in}}. \tag{2}$$

### *B. Graph and graph signal for HCD*

Given two co-registered remote sensing images acquired by different sensors over the same region at different times (*i.e.*, $t_1$, $t_2$), denoted as $\hat{\mathbf{X}} \in \mathbb{R}^{M\times N\times C_x}$ in domain $\mathcal{X}$ and $\hat{\mathbf{Y}} \in \mathbb{R}^{M\times N\times C_y}$ in domain $\mathcal{Y}$, with pixels defined as $\hat{x}(m,n,c)$ and $\hat{y}(m,n,c)$ respectively, the objective of HCD is to find the changed regions represented by a binary map ($CM \in \mathbb{R}^{M\times N}$) that labels changed and unchanged pixels.

Since the heterogeneous images show quite different appearances and characteristics, directly comparing their pixel values is meaningless. The strategy is instead to find the connections between the topological structures of heterogeneous images. We first construct a K-nearest neighbors (KNN) graph for each image.

**Definition 1.** *(KNN Graph). Given a set of data points $\mathbf{z} = \{z_1, z_2, \cdots, z_n\}$, a KNN graph $G = \{\mathcal{V}, \mathcal{E}, \mathbf{W}\}$ consists of $n$ vertices connected by a set of edges $\mathcal{E}$ and their associate weights $\mathbf{W}$, where $\mathcal{V} = \{1, 2, \cdots, n\}$, and $(i,j) \in \mathcal{E}$ if and only if $z_i$ is the KNN of $z_j$ or $z_j$ is the KNN of $z_i$, and $w(i,j)$ is the weight for the edge $(i,j) \in \mathcal{E}$ and zero for others.*

As the scale of a pixel-wise graph that sets each pixel as a vertex is very large (*e.g.*, an image with a size of $500 \times 500$ requires $2.5 \times 10^5$ vertices), we choose the patch-wise or superpixel-wise KNN graph to reduce the computational cost while incorporating the contextual information for each vertex.

For the patch-wise KNN graph $G_{t1}$ of the pre-event image $\hat{\mathbf{X}}$ [42], we first divide the image $\hat{\mathbf{X}}$ into a number of non-overlapped square patches with the size $p \times p \times C_x$, and then vectorize and stack these patches into a patch group matrix $\mathbf{X} \in \mathbb{R}^{N\times M_x}$, where $M_x = p^2 C_x$ and $N$ is the total number of patches. Then $G_{t1}$ can be constructed by setting each patch as a vertex with $\mathcal{V}_{t1} = \mathcal{I}$, $\mathcal{E}_{t1} = \{(i,j)\,|i \in \mathcal{I}; j \in \mathcal{N}_i^{\mathbf{x}}\}$, where $\mathcal{I} = \{1, 2, \cdots, N\}$, $\mathcal{N}_i^{\mathbf{x}} = \mathcal{N}_{in}^{\mathbf{x}} \cup \mathcal{N}_{out}^{\mathbf{x}}$ with

$$\begin{aligned} \mathcal{N}_{in}^{\mathbf{x}} &= \{j | j \in \mathcal{I};\ \mathbf{X}_i \text{ is the KNN of } \mathbf{X}_j\}, \\ \mathcal{N}_{out}^{\mathbf{x}} &= \{j | j \in \mathcal{I};\ \mathbf{X}_j \text{ is the KNN of } \mathbf{X}_i\}. \end{aligned} \tag{3}$$

For the superpixel-wise KNN graph $G_{t1}$ [36], we first segment the images $\hat{\mathbf{X}}$ and $\hat{\mathbf{Y}}$ independently with the simple linear iterative clustering (SLIC) method [51], and then combine the segmentation maps from $\hat{\mathbf{X}}$ and $\hat{\mathbf{Y}}$ though the intersection operator to obtain the co-segmentation map $\mathbf{\Omega} = \{\Omega_i | 1, \cdots, N\}$, which consists of $N$ co-segmented superpixels of $\hat{\mathbf{X}}$ and $\hat{\mathbf{Y}}$, defined as $\hat{\mathbf{X}}_i = \{\hat{x}(m,n,c)\,|\,(m,n) \in \Omega_i, c = 1, \cdots, C_x\}$ and $\hat{\mathbf{Y}}_i = \{\hat{y}(m,n,c)\,|\,(m,n) \in \Omega_i, c = 1, \cdots, C_y\}$, respectively. Then, we extract $M_x$ features (denoted as $\mathbf{X}_i \in \mathbb{R}^M$ ) for superpixel $\hat{\mathbf{X}}_i$, and stack these feature vectors to obtain the feature matrix $\mathbf{X} \in \mathbb{R}^{N\times M}$ . Following that, $G_{t1}$ can be constructed by setting each superpixel as a vertex with $\mathcal{V}_{t1} = \mathcal{I}$ and $\mathcal{E}_{t1} = \{(i,j)\,|i \in \mathcal{I}; j \in \mathcal{N}_i^{\mathbf{x}}\}$.

For the post-event image $\hat{\mathbf{Y}}$, we can construct the patch-wise or superpixel-wise graph $G_{t2}$ in a similar way as $G_{t1}$. Thus, the $i$-th vertex in $G_{t1}$ and the $i$-th vertex in $G_{t2}$ correspond to the same geographical location.

Once the graphs ($G_{t1}$, $G_{t2}$) are constructed to capture the structure information of the heterogeneous images, we can obtain the corresponding graph signals of $\mathbf{X} = \{\mathbf{X}_1, \mathbf{X}_2, \cdots, \mathbf{X}_N\}$ and $\mathbf{Y} = \{\mathbf{Y}_1, \mathbf{Y}_2, \cdots, \mathbf{Y}_N\}$.

### *C. HCD problem on the graph*

Once the graphs and graph signals are constructed, the changes between heterogeneous images can be characterized in two ways: the structure difference between the graphs of $G_{t1}$ and $G_{t2}$, and the signal difference between $\mathbf{X}$ and $\mathbf{Y}$ on the graphs, as illustrated in Fig. 2. However, directly comparing $G_{t1}$, $G_{t2}$ or $\mathbf{X}$, $\mathbf{Y}$ is difficult, as they are constructed on different domains. To avoid the leakage of heterogeneous data, we measure the difference by comparing the responses of the signals (*i.e.*, $\mathbf{X}$, $\mathbf{Y}$) on systems defined on the graphs (*i.e.*, $G_{t1}$, $G_{t2}$).

In the GSP, each vertex diffuses its information to its neighbors and also receives the information diffused from other vertices. Therefore, the GSP is a very effective tool not only for analyzing the signal but also for understanding the structure of graph.

With $\mathbf{X}$ and $\mathbf{Y}$ representing the graph signals, $H(\mathbf{S}_{t1})$ and $H(\mathbf{S}_{t2})$ denoting the graph filters defined on the graphs of $G_{t1}$ and $G_{t2}$ respectively, we have two strategies to measure the changes between heterogeneous images: first, calculating the structure difference between $G_{t1}$ and $G_{t2}$ by comparing the output signals of the same input signal on different graph filters (*e.g.*, comparing the Fig. 2(a) and 2(b), or the Fig. 2(c) and 2(d)), *i.e.*, the differences between $H(\mathbf{S}_{t1})\,\mathbf{X}$ and $H(\mathbf{S}_{t2})\,\mathbf{X}$, or $H(\mathbf{S}_{t1})\,\mathbf{Y}$ and $H(\mathbf{S}_{t2})\,\mathbf{Y}$. Second, calculating the signal difference between $\mathbf{X}$ and $\mathbf{Y}$ by comparing the output signals of the different input signals on the same graph filter (*e.g.*, comparing the Fig. 2(a) and 2(c), or the Fig. 2(b) and 2(d)), *i.e.*, the differences between $H(\mathbf{S}_{t1})\,\mathbf{X}$ and $H(\mathbf{S}_{t1})\,\mathbf{Y}$, or $H(\mathbf{S}_{t2})\,\mathbf{X}$ and $H(\mathbf{S}_{t2})\,\mathbf{Y}$, which will be analyzed in the spectral domain in part II [45].

### *D. Responses of signals on different systems*

We first consider the output of each signal on the filter defined on its own graph. Define the average weight matrix as $\mathbf{W}^{\text{avg}} \triangleq \mathbf{D}^{-1}\mathbf{A}$ with $\mathbf{D}$ representing the degree matrix. By taking $\mathbf{X}$ and $H(\mathbf{S}_{t1})$ as an example, with the simplest case of $H(\mathbf{S}_{t1}) = \mathbf{W}_{t1}^{\text{avg}}$, we have

$$(\mathbf{W}_{t1}^{\text{avg}}\mathbf{X})_i = \frac{1}{|\mathcal{N}_i^{\mathbf{x}}|} \sum_{j \in \mathcal{N}_i^{\mathbf{x}}} \mathbf{X}_j. \tag{4}$$

This filter of $(\mathbf{W}_{t1}^{\text{avg}}\mathbf{X})_i$ characterizes the concentration of information from the neighboring vertices of the understudied vertex $\mathbf{X}_i$. Since $\mathbf{X}_j$ is the KNN of $\mathbf{X}_i$ or $\mathbf{X}_i$ is the KNN of $\mathbf{X}_j$ when $j \in \mathcal{N}_i^{\mathbf{x}}$, we have that $\mathbf{X}_j$ and $\mathbf{X}_i$ are very similar, then we have $(\mathbf{W}_{t1}^{\text{avg}}\mathbf{X})_i \approx \mathbf{X}_i$. This filter of $(\mathbf{W}_{t1}^{\text{avg}}\mathbf{X})_i$ can be regarded as an average smooth operator, as illustrated by the Fig. 3(a) and 3(b).

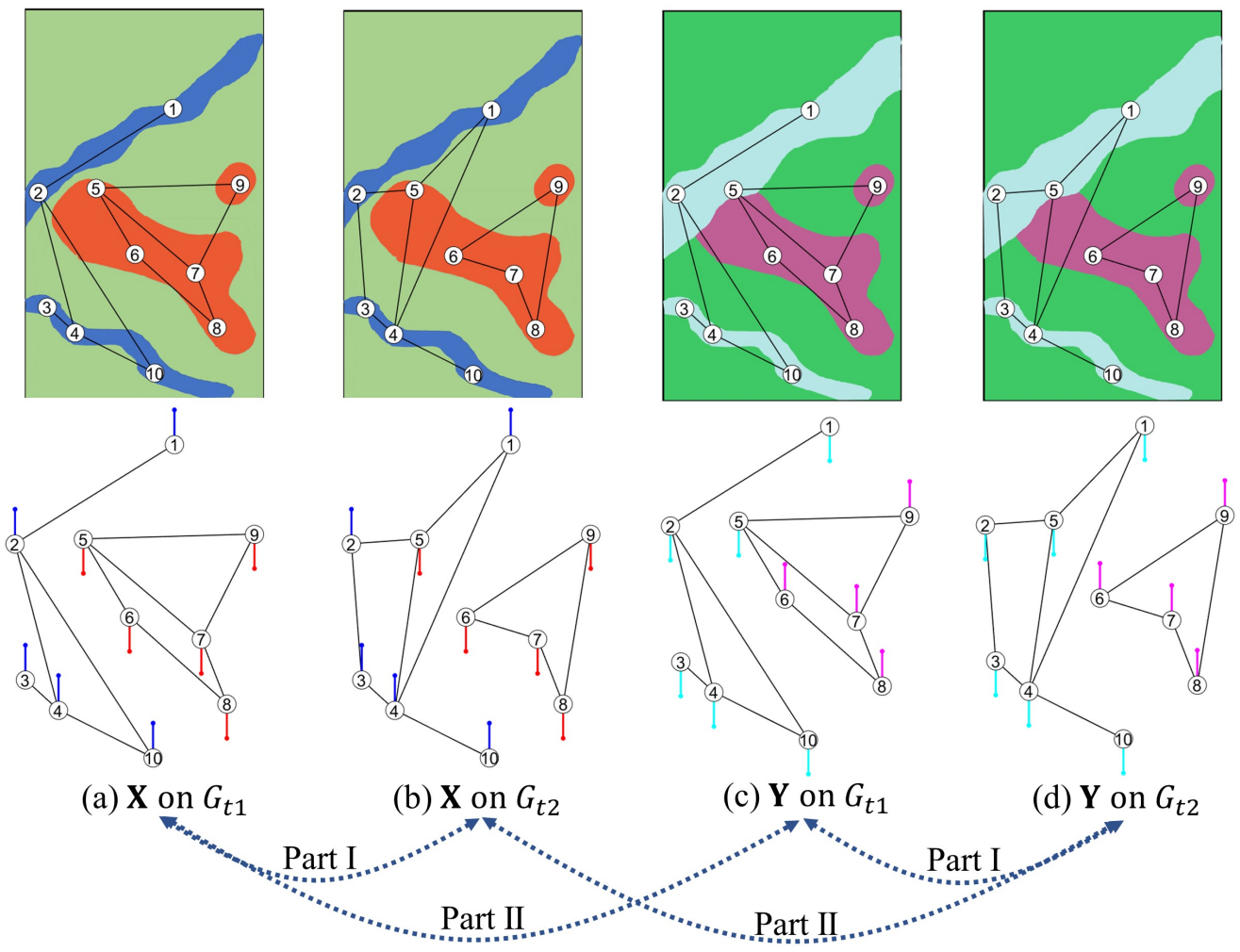

Fig. 2. Two strategies for the HCD from GSP perspective: calculating the structure difference by letting the same graph signal pass through different graph filters (part I); and calculating the signal difference by letting different graph signals pass through the same graph filter (part II).

For the case of $H(\mathbf{S}_{t1}) = \mathbf{P}_{t1}$, we have

$$(\mathbf{P}_{t1}\mathbf{X})_i = \frac{1}{\sum_{j=1}^{N} w_{i,j}^{t1}} \sum_{j=1}^{N} w_{i,j}^{t1}\mathbf{X}_j, \tag{5}$$

where $w_{i,j}^{t1}$ is the $(i,j)$-th element of $\mathbf{W}_{t1}$. Easily, we have $(\mathbf{P}_{t1}\mathbf{X})_i \approx \mathbf{X}_i$. This filter of $(\mathbf{P}_{t1}\mathbf{X})_i$ can be regarded as a weighted smooth operator. By comparing (4) and (5), we can find that $\mathbf{W}_{t1}^{\text{avg}}\mathbf{X}$ is a special case of $\mathbf{P}_{t1}\mathbf{X}$, due to that the adjacent matrix $\mathbf{A}$ can be considered as a special case of the weight matrix $\mathbf{W}$, whereby all nonzero weights are equal to unity.

For the case of $H(\mathbf{S}_{t1}) = \mathbf{L}_{t1}^{\text{rw}}$, we have

$$(\mathbf{L}_{t1}^{\text{rw}}\mathbf{X})_i = \frac{1}{\sum_{j=1}^{N} w_{i,j}^{t1}} \sum_{j=1}^{N} w_{i,j}^{t1}(\mathbf{X}_i - \mathbf{X}_j). \tag{6}$$

Easily, we have $(\mathbf{L}_{t1}^{\text{rw}}\mathbf{X})_i = \mathbf{X}_i - (\mathbf{P}_{t1}\mathbf{X})_i \approx \mathbf{0}$. The filter of $(\mathbf{L}_{t1}^{\text{rw}}\mathbf{X})_i$ characterizes the differences between the signal value on the $i$-th vertex and the signal values on its neighboring vertices, which can be regarded as a difference operator.

Second, we consider the output of each signal on the filter defined on the other graph, and take $\mathbf{X}$ and $H(\mathbf{S}_{t2})$ as an example. Similarly to (4), (5), (6), we have

$$\begin{aligned}
(\mathbf{W}_{t2}^{\text{avg}}\mathbf{X})_i &= \frac{1}{|\mathcal{N}_i^{\mathbf{y}}|} \sum_{j\in\mathcal{N}_i^{\mathbf{y}}} \mathbf{X}_j, \\
(\mathbf{P}_{t2}\mathbf{X})_i &= \frac{1}{\sum_{j=1}^{N} w_{i,j}^{t2}} \sum_{j=1}^{N} w_{i,j}^{t2}\mathbf{X}_j, \\
(\mathbf{L}_{t2}^{\text{rw}}\mathbf{X})_i &= \frac{1}{\sum_{j=1}^{N} w_{i,j}^{t2}} \sum_{j=1}^{N} w_{i,j}^{t2}(\mathbf{X}_i - \mathbf{X}_j).
\end{aligned} \tag{7}$$

Third, we compare the output signals of the same signal on different graph filters. By taking $\mathbf{X}$ and $H(\mathbf{S}) = \mathbf{W}^{\text{avg}}$ as an example, we have

$$\begin{aligned}
\mathbf{d}_i^{\mathbf{x}} &= (\mathbf{W}_{t2}^{\text{avg}}\mathbf{X})_i - (\mathbf{W}_{t1}^{\text{avg}}\mathbf{X})_i \\
&= \frac{1}{|\mathcal{N}_i^{\mathbf{y}}|} \sum_{j'\in\mathcal{N}_i^{\mathbf{y}}} \mathbf{X}_{j'} - \frac{1}{|\mathcal{N}_i^{\mathbf{x}}|} \sum_{j\in\mathcal{N}_i^{\mathbf{x}}} \mathbf{X}_j.
\end{aligned} \tag{8}$$

Intuitively, $\mathbf{d}_i^{\mathbf{x}}$ measures the difference of signal values concentrated at the $i$-th vertex of different systems, which is related to the local structures of two graphs at the $i$-th vertex. Considering the signals of $(\mathbf{X}_i, \mathbf{X}_j)$ and $(\mathbf{Y}_i, \mathbf{Y}_{j'})$ connected by edges of $G_{t1}$ and $G_{t2}$ respectively, and assuming that the regions represented by the $j$-th and $j'$-th vertices are unchanged, then we have $\mathbf{X}_i$ and $\mathbf{X}_j$ are very similar (representing the same kind of object), $\mathbf{Y}_i$ and $\mathbf{Y}_{j'}$ are also very similar (representing the same kind of object), and: 1) if the $i$-th vertex is unchanged in the event, then $\mathbf{X}_j$ and $\mathbf{X}_{j'}$ also represent the same kind of object (showing that $\mathbf{X}_j$ and $\mathbf{X}_{j'}$ are also very similar), which makes the elements of $\mathbf{d}_i^{\mathbf{x}}$ very small; 2) if the $i$-th vertex is changed in the event, then $\mathbf{X}_j$ and $\mathbf{X}_{j'}$ represent the different kinds of object (showing that $\mathbf{X}_j$ and $\mathbf{X}_{j'}$ are different), which makes the elements of $\mathbf{d}_i^{\mathbf{x}}$ large. Therefore, we can find that the $\mathbf{d}_i^{\mathbf{x}}$ can be used to measure the change probability (level) of the $i$-th vertex, as illustrated by the Fig. 3(a) and Fig. 4(a).

For the $H(\mathbf{S}) = \mathbf{P}$, we similarly have

$$\begin{aligned}
\mathbf{d}_i^{\mathbf{x}} &= (\mathbf{P}_{t2}\mathbf{X})_i - (\mathbf{P}_{t1}\mathbf{X})_i \\
&= \sum_{j=1}^{N} \left( \frac{w_{i,j}^{t2}\mathbf{X}_j}{\sum_{j=1}^{N} w_{i,j}^{t2}} - \frac{w_{i,j}^{t1}\mathbf{X}_j}{\sum_{j=1}^{N} w_{i,j}^{t1}} \right).
\end{aligned} \tag{9}$$

Since and $\mathbf{L}^{\text{rw}} = \mathbf{I} - \mathbf{P}$, for the $H(\mathbf{S}) = \mathbf{L}^{\text{rw}}$, we have

$$\mathbf{d}_i^{\mathbf{x}} = (\mathbf{L}_{t2}^{\text{rw}}\mathbf{X})_i - (\mathbf{L}_{t1}^{\text{rw}}\mathbf{X})_i = (\mathbf{P}_{t1}\mathbf{X})_i - (\mathbf{P}_{t2}\mathbf{X})_i. \tag{10}$$

It should be noted that when $\mathbf{W}$ is not normalized by rows (*i.e.*, $\mathbf{W}\mathbf{1}_N \neq c\mathbf{1}_N$ with $c$ being a non-zero constant), $H(\mathbf{S}) = \mathbf{W}$ is not recommended for calculating the changes, *i.e.*, $\mathbf{d}_i^{\mathbf{x}} = (\mathbf{W}_{t2}\mathbf{X})_i - (\mathbf{W}_{t1}\mathbf{X})_i$ is not appropriate with $\mathbf{W}$ that is unnormalized by rows. This is because it will cause the leakage of heterogeneous data. For example, if we assume $\mathbf{Y} = a\mathbf{X} + b\mathbf{I}$ with $a$ and $b$ being non-zero constants (indicating no change between images) and set $w_{i,j}^{t1} = \frac{1}{\|\mathbf{X}_i - \mathbf{X}_j\|_2}$, $w_{i,j}^{t2} = \frac{1}{\|\mathbf{Y}_i - \mathbf{Y}_j\|_2}$, then we have $\mathbf{W}_{t2}\mathbf{X} - \mathbf{W}_{t1}\mathbf{X} = \frac{1-a}{a}\mathbf{W}_{t1}\mathbf{X}$, which can not indicate changes. To avoid the leakage and confusion of heterogeneous data, we use the graph shift operator of normalized average weighting matrix $\mathbf{W}^{\text{avg}}$, or the normalized $\mathbf{P}$ and $\mathbf{L}^{\text{rw}}$ for calculating the $\mathbf{d}_i^{\mathbf{x}}$ in (8), (9), (10).

### *E. Higher-order operators*

Furthermore, we only consider the simplest cases of $H(\mathbf{S}) = \mathbf{W}^{\text{avg}}, \mathbf{P}, \mathbf{L}^{\text{rw}}$ in (8), (9), (10). We can also choose $H(\mathbf{S})$ as the polynomials of these operators, *i.e.*, $H(\mathbf{S}) = \sum_{m=1}^{M} h_m \mathbf{S}^m$. Then, we can rewrite (8), (9), (10) as

$$\begin{aligned}
\mathbf{d}_i^{\mathbf{x}} &= (H(\mathbf{S}_{t2})\mathbf{X})_i - (H(\mathbf{S}_{t1})\mathbf{X})_i \\
&= \sum_{j=1}^{N} \sum_{m=1}^{M} (h_m \mathbf{S}_{t2}^m - h_m \mathbf{S}_{t1}^m)\mathbf{X}_j.
\end{aligned} \tag{11}$$

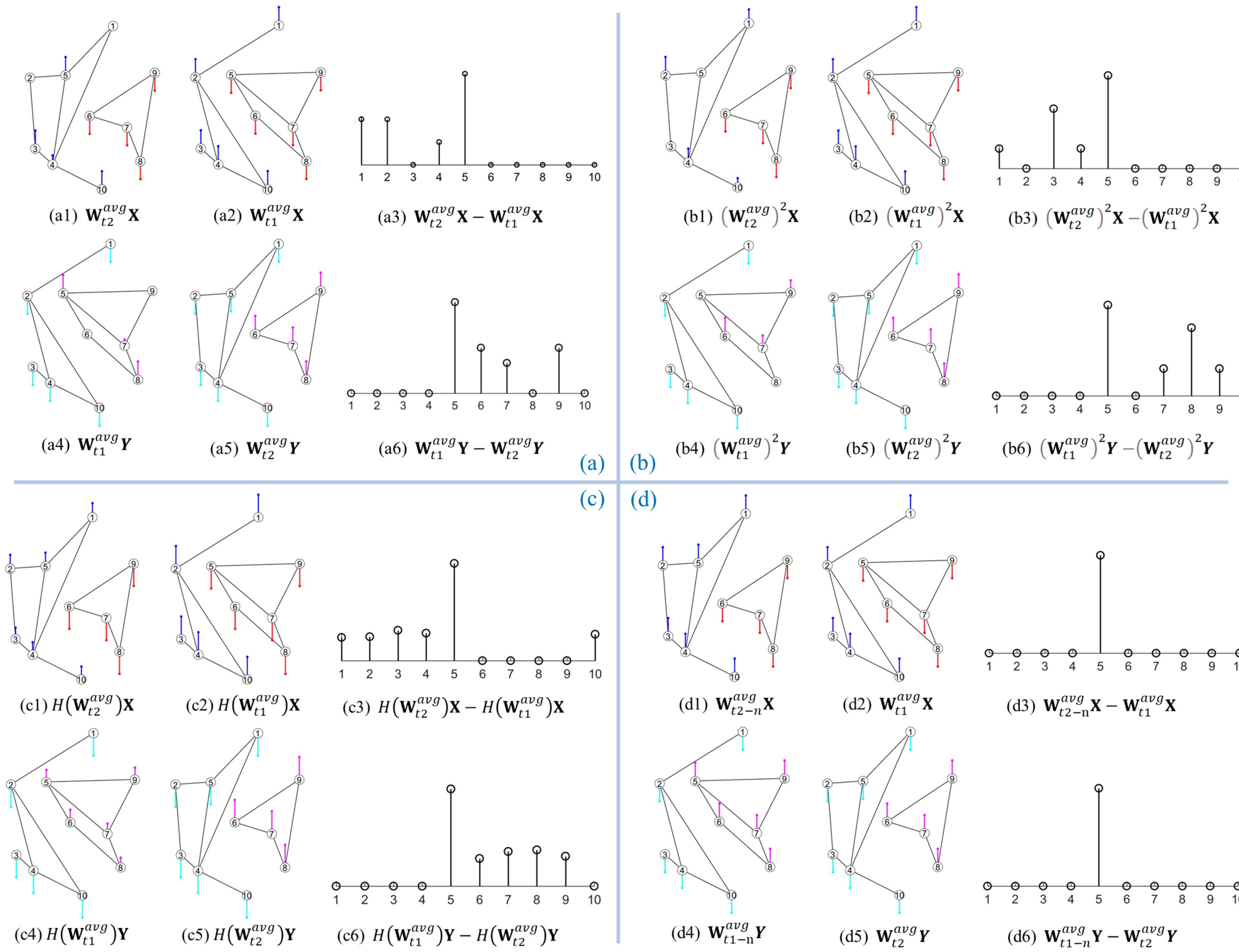

Fig. 3. The change measurements with different forms, with the graphs and graph signals from Fig. 2 and the filters in Fig. 4. (a) Measure the changes using (8), where $\mathbf{W}_{t1}^{\text{avg}}\mathbf{X}=\mathbf{X}$ and $\mathbf{W}_{t2}^{\text{avg}}\mathbf{Y}=\mathbf{Y}$, $\left(\mathbf{W}_{t2}^{\text{avg}}\mathbf{X}\right)_5$ differs from $\mathbf{X}_5$, and $\left(\mathbf{W}_{t1}^{\text{avg}}\mathbf{Y}\right)_5$ differs from $\mathbf{Y}_5$ due to the structure changes. (b) Measure the changes using the second order information, where the structure changes between $G_{t1}$ and $G_{t2}$ can also be reflected. (c) Measure the changes using (11), where the filter $H(\mathbf{S})$ uses the form of Fig. 4c to make its transfer function approximate a low-pass filter. The structure difference between $G_{t1}$ and $G_{t2}$ shows up more clearly in Fig. 3c than in Fig. 3a-3b. (d) Measure the changes by reducing the influence of changed vertex, where the information propagation from the changed vertex to its neighboring vertices is cut off, as illustrated by the $\mathbf{W}_{t1-n}^{\text{avg}}$ and $\mathbf{W}_{t2-n}^{\text{avg}}$ in Fig. 4d.

Equation (11) can be interpreted in two ways: 1) $\left(H(\mathbf{S}_{t1})\mathbf{X}\right)_i=\sum_{m=1}^{M}h_m(\mathbf{S}_{t1}^m\mathbf{X})_i$ is a weighted sum of the attributes of vertices that are within $M$-hop away from the $i$-th vertex. The coefficient $h_m$ quantifies the contribution from the $m$th-hop neighbors. In this way, we can treat the $H(\mathbf{S}_{t1})$ as a new operator of weighted higher-order graph that explores the high-order neighborhood information of $G_{t1}$. Then, $\mathbf{d}^{\mathbf{x}}=\left(\sum_{m=1}^{M}h_m\mathbf{S}_{t2}^m-\sum_{m=1}^{M}h_m\mathbf{S}_{t1}^m\right)\mathbf{X}$ is the difference between responses of $\mathbf{X}$ on different higher-order graphs. 2) $\mathbf{S}_{t1}^m\mathbf{X}=\mathbf{S}_{t1}\left(\mathbf{S}_{t1}^{m-1}\mathbf{X}\right)$ can be regarded as a high-order filtering process that repeats operator $\mathbf{S}_{t1}$ by $m$ times. Then, $\mathbf{d}^{\mathbf{x}}=\sum_{m=1}^{M}h_m\left(\mathbf{S}_{t2}^m\mathbf{X}-\mathbf{S}_{t1}^m\mathbf{X}\right)$ is the weighted sum of differences between responses of $\mathbf{X}$ on the filters of different orders.

Note that our aim is to discover changes in the structures between graphs ($G_{t1}$ and $G_{t2}$) by comparing the responses of the same signal $\mathbf{X}$ on different filters $H(\mathbf{S})$, then the DI using this type of filters $H(\mathbf{S})=\sum_{m=1}^{M}h_m\mathbf{S}^m$ is able to fully exploit information about changes in graph structures, both in terms of weighted higher-order graph and in terms of weighted higher-order filtering, as illustrated by the Fig. 3(b)-3(c) and Fig. 4(b)-4(c).

### F. The influence of changes

In the analysis of (8), we have assumed that the neighboring vertex of the given $i$-th vertex is unchanged. This assumption is reasonable due to the typical sparse prior of CD, that is, only a small part of the area changes and most areas remains unchanged during the event in practice. However, these changed vertices have a negative impact on the CD results. Next, we will show this influence from the view of signal propagation.

As shown in Fig. 3(a), for the $i$-th vertex, if one of its KNN in $G_{t2}$ is changed during the event, *e.g.*, the $j'$-th vertex ($j'\in\mathcal{N}_i^{\mathbf{y}}$) is changed, we have that this changed signal will propagate to the $i$-th vertex by the graph shift operator, such as $\mathbf{W}_{t2}^{\text{avg}}\mathbf{X}$. 1) If the $i$-th vertex is unchanged, this propagated signal $\mathbf{X}_{j'}$ differs significantly from $\mathbf{X}_i$, which increases the difference between $(\mathbf{W}_{t2}^{\text{avg}}\mathbf{X})_i$ and $(\mathbf{W}_{t1}^{\text{avg}}\mathbf{X})_i$ for the unchanged $i$-th vertex. For example, the unchanged vertices 1, 2 and 4 connected with the changed vertex 5 in $G_{t2}$ are affected by the changed vertex, as illustrated by Fig. 3(a3). 2) If the $i$-th vertex is changed and the $j'$-th vertex changed to the same category as the $i$-th vertex, then this propagated signal $\mathbf{X}_{j'}$ is quite similar as the $\mathbf{X}_i$, which reduce the difference between $\left(\mathbf{W}_{t2}^{\text{avg}}\mathbf{X}\right)_i$ and $\left(\mathbf{W}_{t1}^{\text{avg}}\mathbf{X}\right)_i$ for the changed $i$-th vertex. From the above analysis, we can find that the changed vertex will affect the judgement of the state of other vertices whose connected neighbors contain this change vertex, *i.e.*, making the $\mathbf{d}^{\mathbf{x}}$ less discriminative.

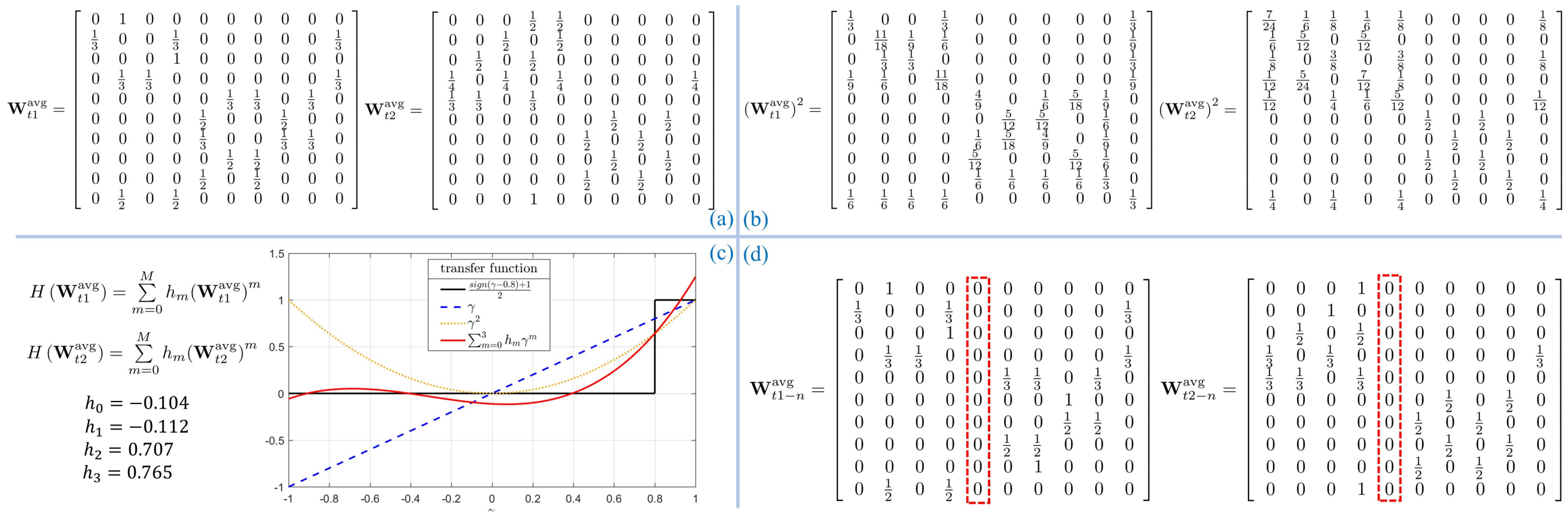

Fig. 4. Different filters used in change measurements of Fig. 3 with graphs from Fig. 2. (a) $\mathbf{W}_{t1}^{\text{avg}}$ and $\mathbf{W}_{t2}^{\text{avg}}$. (b) $\left(\mathbf{W}_{t1}^{\text{avg}}\right)^2$ and $\left(\mathbf{W}_{t2}^{\text{avg}}\right)^2$. (c) $H\left(\mathbf{W}_{t1}^{\text{avg}}\right)$ and $H\left(\mathbf{W}_{t2}^{\text{avg}}\right)$, and the transfer functions of the filters in (a)-(c). $H\left(\mathbf{W}_{t1}^{\text{avg}}\right)$ and $H\left(\mathbf{W}_{t2}^{\text{avg}}\right)$ are approximate low-pass filters with the transfer function $H(\mathbf{\Gamma}) = \frac{sign(\mathbf{\Gamma}-0.8)+1}{2}$, refer to part II for more details of the spectral property and transfer function. (d) $\mathbf{W}_{t1-n}^{\text{avg}}$ and $\mathbf{W}_{t2-n}^{\text{avg}}$, where the information propagation from the changed vertex to its neighboring vertices is cut off.

## III. HCD Method Based on the Vertex Domain Filtering

In this section, we propose a framework for the HCD by using the vertex domain filtering. We choose the $H(\mathbf{S}) = \sum_{m=1}^{M} h_m \mathbf{P}^m$, and rewrite (11) as

$$\begin{aligned} \mathbf{d}_i^{\mathbf{x}} &= \left(H(\mathbf{P}_{t2})\mathbf{X}\right)_i - \left(H(\mathbf{P}_{t1})\mathbf{X}\right)_i \\ &= \sum_{m=1}^{M}\sum_{j=1}^{N} h_m\left((\mathbf{P}_{t2}^m)_{i,j}\mathbf{X}_j - (\mathbf{P}_{t1}^m)_{i,j}\mathbf{X}_j\right) \\ &= \sum_{m=1}^{M}\sum_{j=1}^{N} h_m\left((\mathbf{P}_{t2}^m)_{i,j}\mathbf{X}_j - (\mathbf{P}_{t2}^m)_{i,j}\mathbf{X}_i\right. \\ &\qquad \left. + (\mathbf{P}_{t1}^m)_{i,j}\mathbf{X}_i - (\mathbf{P}_{t1}^m)_{i,j}\mathbf{X}_j\right) \\ &= \sum_{m=1}^{M}\sum_{j=1}^{N} h_m\left((\mathbf{P}_{t2}^m)_{i,j} - (\mathbf{P}_{t1}^m)_{i,j}\right)(\mathbf{X}_i - \mathbf{X}_j), \end{aligned} \tag{12}$$

where the second equality comes from $\mathbf{P}\mathbf{1}_N = \mathbf{1}_N$. If we choose $\mathbf{S} = \mathbf{W}^{\text{avg}}$ or $\mathbf{S} = \mathbf{L}^{\text{rw}}$, we also have

$$\mathbf{d}_i^{\mathbf{x}} = \sum_{m=1}^{M}\sum_{j=1} h_m\left((\mathbf{S}_{t2}^m)_{i,j} - (\mathbf{S}_{t1}^m)_{i,j}\right)(\mathbf{X}_i - \mathbf{X}_j), \tag{13}$$

with the equations of $\mathbf{W}^{\text{avg}}\mathbf{1}_N = \mathbf{1}_N$ and $\mathbf{L}^{\text{rw}}\mathbf{1}_N = \mathbf{0}$.

Equation (13) provides the change features for each vertex, which can be regarded as the difference after the concentration of the signal difference ($\Delta_{i,j}^{\mathbf{x}} = \mathbf{X}_i - \mathbf{X}_j$) at each vertex on graphs $G_{t1}$ and $G_{t2}$. In order to obtain the change level of each vertex, we change the signal difference $\Delta_{i,j}^{\mathbf{x}}$ as the signal distance $dist_{i,j}^{\mathbf{x}} = \|\mathbf{X}_i - \mathbf{X}_j\|_2^2$, and calculate the change level as

$$\begin{aligned} f_i^{\mathbf{x}} &= \sum_{m=1}^{M}\sum_{j=1}^{N} h_m\left((\mathbf{S}_{t2}^m)_{i,j} - (\mathbf{S}_{t1}^m)_{i,j}\right) dist_{i,j}^{\mathbf{x}} \\ &= \sum_{j=1}^{N}\sum_{m=1}^{M} h_m((\mathbf{S}_{t2}^m - \mathbf{S}_{t1}^m) \odot dist^{\mathbf{x}})_{i,j}, \end{aligned} \tag{14}$$

where $dist^{\mathbf{x}}$ represents the distance matrix of pre-event image and $\odot$ denotes the Hadamard product.

In (14), we only calculate the forward change level $\mathbf{f}^{\mathbf{x}}$ that comparing the outputs of $\mathbf{X}$ on the graph filters of $H(\mathbf{S}_{t1})$ and $H(\mathbf{S}_{t2})$. Similarly, we can obtain the backward change level $\mathbf{f}^{\mathbf{y}}$ by comparing the outputs of $\mathbf{Y}$ on the graph filters of $H(\mathbf{S}_{t1})$ and $H(\mathbf{S}_{t2})$, that is

$$\mathbf{f}^{\mathbf{y}} = \left(\sum_{m=1}^{M} h_m\left(\mathbf{S}_{t1}^m - \mathbf{S}_{t2}^m\right) \odot dist^{\mathbf{y}}\right)\mathbf{1}_N, \tag{15}$$

where $dist^{\mathbf{y}}$ represents the distance matrix of post-event image with element being $dist_{i,j}^{\mathbf{y}} = \|\mathbf{Y}_i - \mathbf{Y}_j\|_2^2$.

### A. Change Extraction

Once the forward and backward DIs ($\mathbf{f}^{\mathbf{x}}$ and $\mathbf{f}^{\mathbf{y}}$) are obtained, the CM solution can be regarded as a binary segmentation problem, which can be solved by some thresholding methods (such as Otsu [52]) or clustering methods (such as K-means [53], fuzzy c-means clustering [54]) as used in homogeneous CD. Because the main purpose of this paper is to propose a GSP perspective for HCD problem, the segmentation method for computing the final CM is not our focus, while may also conceal the contribution of this paper. Therefore, we directly choose the Markov random field (MRF) co-segmentation method [36] to label the vertices, which can fuse the forward and backward DIs in the segmentation process.

TABLE II
THE OVERALL FRAMEWORK OF VDF-HCD.

Algorithm 1. VDF-HCD

**Input:** Images of $\hat{\mathbf{X}}$ and $\hat{\mathbf{Y}}$, parameters of $N$, $Iter$.
Choose the operator $\mathbf{S}$ and coefficients $h_m$ to obtain filter $H(\mathbf{S})$.
**Preprocessing:** Construct the graph and graph signals
  Divide the images into patches or superpixels.
  Construct the graph signals of $\mathbf{X}$ and $\mathbf{Y}$.
  Construct the KNN graphs of $G_{t1}$ and $G_{t2}$.
**Main iteration loop of VDF-HCD:**
  Set initial index subset as $\mathcal{S}^0 = \mathcal{I}$.
  for $i = 1, 2, \cdots, Iter$ do
    Construct the new graphs of $G_{t1-n}$ and $G_{t2-n}$ with $\mathcal{S}^{i-1}$.
    Calculate the change levels $\mathbf{f}^{\mathbf{x}}$ and $\mathbf{f}^{\mathbf{y}}$ as
    $\mathbf{f}^{\mathbf{x}} = \left(\sum_{m=1}^{M} h_m \left(\mathbf{S}_{t2-n}^m - \mathbf{S}_{t1}^m\right) \odot dist^{\mathbf{x}}\right) \mathbf{1}_N$
    $\mathbf{f}^{\mathbf{y}} = \left(\sum_{m=1}^{M} h_m \left(\mathbf{S}_{t1-n}^m - \mathbf{S}_{t2}^m\right) \odot dist^{\mathbf{y}}\right) \mathbf{1}_N$
    Segment the DI to obtain the $\mathcal{S}^i$ and $\mathcal{T}^i$.
  end for
**Output:** Compute the final change map with $\mathcal{S}$ and $\mathcal{T}$.

Then, we can obtain the unchanged index subset $\mathcal{S}$ and changed index subset $\mathcal{T}$ of vertices.

### B. Reducing the impact of changes

In subsection II-F, we have analyzed the negative effects of changed vertex on the change measurement from the view of signal propagation. In order to reduce this negative influence, we need to avoid propagating the signal on the changed vertex (*e.g.*, the $j$-th vertex) to their neighboring vertices ($\{i | i \in \mathcal{N}_j^{\mathbf{y}}\}$) by setting $(\mathbf{A}_{t2})_{i,j} = 0$, *i.e.*, connecting each vertex with its neighbors in the unchanged index subset $\mathcal{S}$. However, we cannot identify in advance which vertices are changed, so we employ an iterative framework to complete the elimination of changed vertices in the graph construction. That is, we propagate the unchanged vertices $\mathcal{S}$ computed by the MRF co-segmentation of previous round back to the graph construction process of next round, to obtain the new graphs of $G_{t1-n}$ and $G_{t2-n}$. Then, we can compare the $H(\mathbf{S}_{t2-n})\mathbf{X}$ and $H(\mathbf{S}_{t1})\mathbf{X}$, or $H(\mathbf{S}_{t1-n})\mathbf{Y}$ and $H(\mathbf{S}_{t2})\mathbf{Y}$, to enhance the distinction between changed and unchanged classes in the DI calculation. As illustrated by Fig. 3(d), the structure difference between $G_{t1}$ and $G_{t2}$ shows up most clearly by using the filters in Fig. 4(d).

The overall framework of the proposed vertex domain filtering based HCD method (VDF-HCD) is summarized in Table II.

## IV. DISCUSSIONS

### A. Connection with some other graph based methods

In INLPG [42] and IRG-McS [36], the patch-wise graphs and superpixel-wise graphs are constructed for heterogeneous images respectively, then the graph projection are used to compute the DIs as follows

$$
\begin{aligned}
d_i^{\mathbf{x}} &= \frac{1}{K}\left(\sum_{j' \in \mathcal{N}_i^{\mathbf{y}}} dist_{i,j'}^{\mathbf{x}} - \sum_{j \in \mathcal{N}_i^{\mathbf{x}}} dist_{i,j}^{\mathbf{x}}\right), \\
d_i^{\mathbf{y}} &= \frac{1}{K}\left(\sum_{j' \in \mathcal{N}_i^{\mathbf{x}}} dist_{i,j'}^{\mathbf{y}} - \sum_{j \in \mathcal{N}_i^{\mathbf{y}}} dist_{i,j}^{\mathbf{y}}\right),
\end{aligned} \tag{16}
$$

which measures the structure difference by how different the two KNN position sets of $\mathcal{N}_i^{\mathbf{x}}$ and $\mathcal{N}_i^{\mathbf{y}}$ are in the images. With the definition of the adjacent matrix $\mathbf{A}$ of KNN graph, (16) can be rewritten as

$$
\begin{aligned}
d_i^{\mathbf{x}} &= \frac{1}{K}\sum_{j=1}^{N} \left[(\mathbf{A}_{t2} - \mathbf{A}_{t1}) \odot dist^{\mathbf{x}}\right]_{i,j}, \\
d_i^{\mathbf{y}} &= \frac{1}{K}\sum_{j=1}^{N} \left[(\mathbf{A}_{t1} - \mathbf{A}_{t2}) \odot dist^{\mathbf{y}}\right]_{i,j}.
\end{aligned} \tag{17}
$$

Then, we can find that (17) is a special case of (14), (15) with $\mathbf{S} = \mathbf{W}^{\text{avg}}$ and $m = 1$, that is, INLPG and IRG-McS only consider the first-order information of the graph and ignore the exploitation of higher-order information. At the same time, INLPG and IRG-McS do not take a GSP perspective on HCD issues as this paper does, which leads to a limited application of these methods.

In [39], a point-wise approach based on graph theory is proposed for homogeneous CD of SAR images, which constructs a point-wise graph on a set of characteristic points, then calculates the change level of each vertex as

$$
f_i = \frac{1}{\sum_{j \in \mathcal{N}^{\mathbf{x}}} w_{i,j}^{t1}} \sum_{j \in \mathcal{N}^{\mathbf{x}}} w_{i,j}^{t1} \left|\log \bar{x}_j - \log \bar{y}_j\right|, \tag{18}
$$

where $\bar{x}_j$ and $\bar{y}_j$ represents the mean value of a small patch around the $j$-th vertex in images $\hat{\mathbf{X}}$ and $\hat{\mathbf{Y}}$, respectively. Equation (18) measures the difference of different signal on the same graph, which is similar as $\mathbf{f} = \mathbf{P}_{t1}(\log \mathbf{X} - \log \mathbf{Y})$. However, we can find that (18) cannot be applied to HCD because of the direct comparison between pixels of images, which will cause the leakage of heterogeneous data. At the same time, the high-order information and the influence of the changed vertices are not taken into account in this method.

### B. Extended to homogeneous CD

As mentioned above, the proposed method compares the structures of images, so it is not sensitive to the interference factors in some complicated homogeneous CD problems, such as illumination, season and noise. In addition, it has been demonstrated in INLPG [42] that the structure difference-based operator is more robust to noise than the traditional difference operator (and log-ratio operator) in homogeneous CD of optical images (and SAR images). Therefore, the method proposed in this paper can be directly extended to the homogeneous CD. At the same time, fusing the proposed method with other methods (such as the graph based method in [39]) may further improve the detection performance.

### C. Framework of GSP-VDF based CD

Here, we give a general framework for CD problem (with both homogeneous and heterogeneous multi-temporal remote sensing images) based on the GSP with vertex domain filtering (VDF).

Step 1. Construct the graph and graph signals.

Step 2. Choose the filter $H(\mathbf{S})$ and calculate the DI.

TABLE III
DESCRIPTION OF THE SEVEN HETEROGENEOUS DATA SETS.

| Dataset | Sensor | Size (pixels) | Date | Location | Event (& Spatial resolution) |
|---|---|---|---|---|---|
| #1 | Landsat-5/Google Earth | $300 \times 412 \times 1(3)$ | Sept. 1995 - July 1996 | Sardinia, Italy | Lake expansion (30m.) |
| #2 | Landsat-5/Landsat-8 | $1534 \times 808 \times 7(10)$ | Aug. 2011 - June 2013 | Texas, USA | Forest fire (30m.) |
| #3 | Pleiades/WorldView2 | $2000 \times 2000 \times 3(3)$ | May 2012 - July 2013 | Toulouse, France | Construction (0.52m.) |
| #4 | Spot/NDVI | $990 \times 554 \times 3(1)$ | 1999 - 2000 | Gloucester, England | Flooding ($\approx$ 25m.) |
| #5 | Radarsat-2/Google Earth | $593 \times 921 \times 1(3)$ | June 2008 - Sept. 2012 | Shuguang Village, China | Building construction (8m.) |
| #6 | Landsat-8/Sentinel-1A | $875 \times 500 \times 11(3)$ | Jan. 2017 - Feb. 2017 | Sutter County, USA | Flooding ($\approx$ 15m.) |
| #7 | Radarsat-2/Google Earth | $343 \times 291 \times 1(3)$ | June 2008 - Sept. 2010 | Yellow River, China | Embankment change (8m.) |

Step 3. Segment the DI to obtain the changed and unchanged indices.

Step 4. Repeat steps 1 to 3 to reduce the change influence until the algorithm converges.

In the step 1, it requires that the constructed graph can represent the structure of image and the graph signal can represent the characteristics of the area denoted by each vertex. In the step 2, the $H(\mathbf{S})$ should be able to reflect the differences in signals on different graphs, *i.e.*, be able to reflect the structural properties of the graphs, *e.g.*, low-pass filter for $H(\mathbf{W}^{\text{avg}})$ and $H(\mathbf{P})$, high-pass filter for $H(\mathbf{L}^{\text{rw}})$ according to the analysis in part II [45]. Meanwhile, in the homogeneous CD, the $\mathbf{d}^{\mathbf{x}} = H(\mathbf{S}_{t1})\mathbf{X} - H(\mathbf{S}_{t1})\mathbf{Y}$ and $\mathbf{d}^{\mathbf{y}} = H(\mathbf{S}_{t2})\mathbf{X} - H(\mathbf{S}_{t2})\mathbf{Y}$ can also be fused in the DI calculation. In the step 3, many existing methods can be selected for this binary segmentation problem. At last, the change influence is reduced by an iterative strategy that cuts off the propagation from changed vertices to their neighboring vertices.

## V. EXPERIMENTS

In this section, we analyze the performance of the proposed VDF-HCD. In order to assess the robustness, we use seven pairs of heterogeneous images of varying HCD conditions.

### A. Experimental Setting

*1) Heterogeneous data sets:* First, we introduce the heterogeneous data sets used in this paper (for both part I and part II), as listed in Table III. These data sets contain different types of heterogeneity: multi-sensor image pairs (*e.g.*, #1, #2, #3, #4) and multisource image pairs (*e.g.*, #5, #6, #7), which also show quite different HCD conditions: different resolutions (from 0.52m to 30m), different sizes (from 300 to 2000 pixels in length or width), and different change events (flooding, fire, construction). These heterogeneous data sets are able to evaluate the generalizability and robustness of the proposed method.

*2) Metrics:* We evaluate the detection ability of the methods quantitatively through two types of metrics: first, we assess the DI by the receiver operating characteristic (ROC) and precision-recall (PR) curves, along with the areas under the ROC curve (AUR) and PR curve (AUP) scores. Second, we assess the CM by the overall accuracy (OA), F1-measure (Fm), and Kappa coefficient (Kc), which are calculated by : OA = (TP + TN)/(TP + TN + FP + FN), Fm = (2TP)/(2TP + FP + FN), and Kc = (OA − PRE)/(1 − PRE) with

$$\text{PRE} = \frac{(\text{TP}+\text{FN})(\text{TP}+\text{FP}) + (\text{TN}+\text{FP})(\text{TN}+\text{FN})}{(\text{TP}+\text{TN}+\text{FP}+\text{FN})^2}, \quad (19)$$

where TP, FP, TN, and FN represent the true positives, false positives, true negatives, and false negatives, respectively. Besides, we also provide the detection maps (DI and CM) to evaluate their qualities by visual inspection.

*3) Implementation detail:* We use the supervised-wise KNN graph for VDF-HCD with $N = 5000$ and $K = \sqrt{N}$, and use the mean, median, and variance values of superpixel to calculate the distances of $dist^{\mathbf{x}}$ and $dist^{\mathbf{y}}$ for simplicity. We choose the graph shift operator as $\mathbf{S} = \mathbf{W}^{\text{avg}}$ and set $H(\mathbf{W}^{\text{avg}})$ to obtain an approximate low-pass filter, whose transfer function in the spectral domain is a fourth-order ($M = 4$) polynomial approximation of the truncation function $H(\mathbf{\Gamma}) = \frac{sign(\mathbf{\Gamma}-\gamma_{cf})+1}{2}$ with the cut-off frequency being $\gamma_{cf} = 0.9$, as analyzed in the part II of this paper [45].

### B. Experimental results

*1) Difference images:* First, we show the forward and backward DIs of different data sets generated by VDF-HCD in Fig. 5. We can find that the DIs are able to highlight the changes very well, which demonstrates the effectiveness of the comparison of structures in VDF-HCD. In Fig. 6, we plot the ROC and PR curves of the DIs obtained by VDF-HCD, where the fused DIs are generated by directly summing the forward and backward DIs. The corresponding AUR of ROC curves and AUP of PR curves are listed in Table IV.

From these results, we can see that the VDF-HCD can obtain a better DI in data sets of #2, #4, #5 and #7, which gains a larger AUR and AUP scores as reported in Table IV. Therefore, it is able to obtain a satisfactory CM by directly segmenting the DIs with a simple thresholding method (*e.g.*, Otsu thresholding [52]) or clustering method (*e.g.*, K-means clustering [53]) for Datasets #2, #4, #5 and #7. For the Dataset #3, it contains more types of ground objects than other data sets, *e.g.*, buildings, grass, roads, pitches, and the resolution of the images is very high (0.52m). It means that images of Dataset #3 contain many categories of features, which leads to an increase in the difficulty of HCD task. From Fig. 5 and Table IV, we can find that DI performance of Dataset #3 is not as good as the other data sets because of the complex texture structure of the two images. By comparing the forward DI and backward DI of each data set, it can be observed that they can provide complementary information, which in turn can be fused to obtain better detection results. This complementary advantage of DIs can also be seen visually in Fig. 3, where a more obvious contrast of changed vertex and unchanged vertex can be obtained by directly summing the $\mathbf{d}^{\mathbf{x}}$ and $\mathbf{d}^{\mathbf{y}}$.

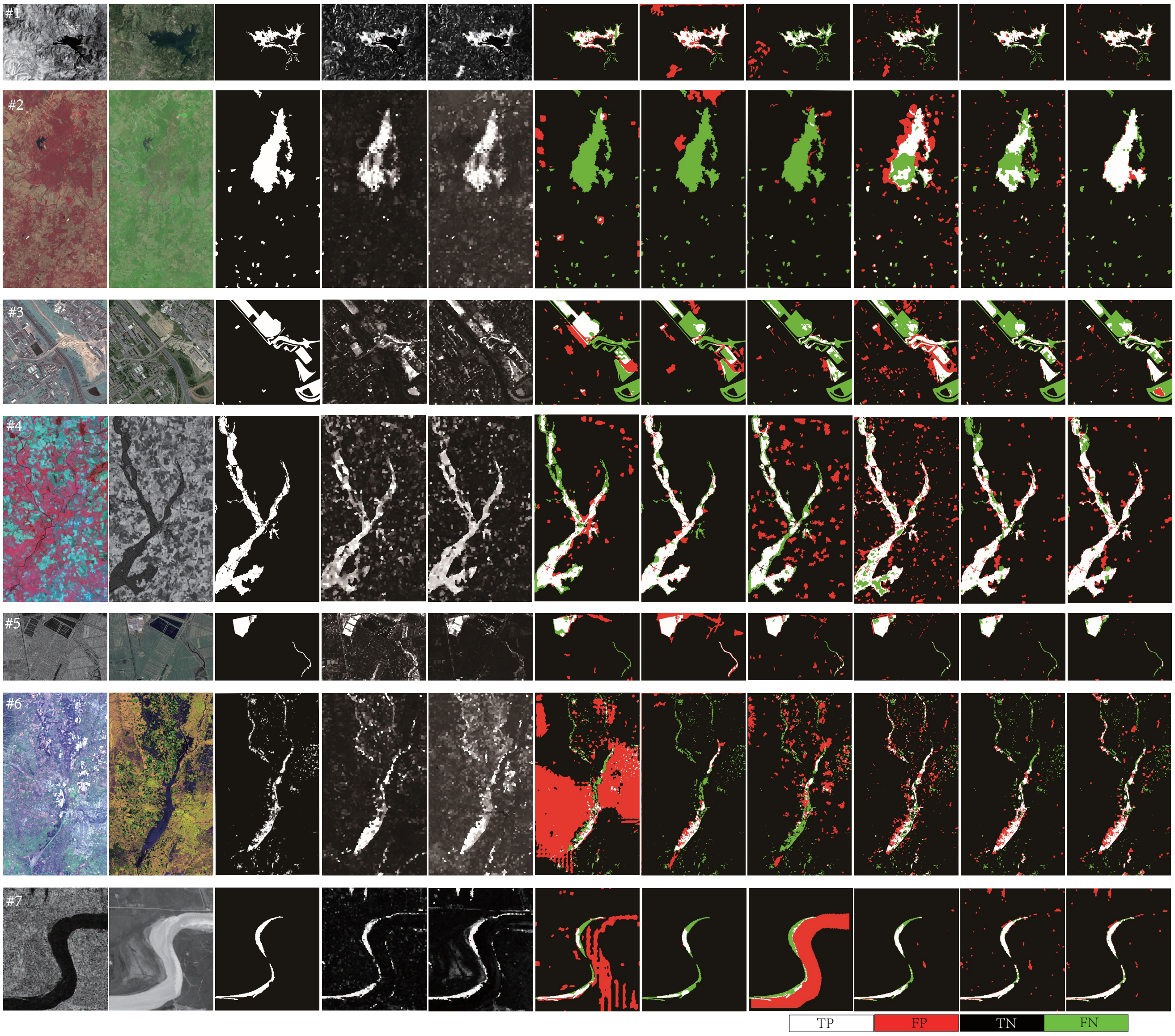

Fig. 5. DI of VDF-HCD and binary CMs of different methods on heterogeneous data sets. From top to bottom, they correspond to Datasets #1 to #7, respectively. From left to right are: (a) pre-event image; (b) post-event image; (c) ground truth; (d) forward DI of VDF-HCD; (e) backward DI of VDF-HCD; (f) binary CM of M3CD; (g) binary CM of FPMS; (h) binary CM of CICM; (i) binary CM of NPSG; (j) binary CM of IRG-McS; (k) binary CM of VDF-HCD. In the binary CM, White: true positives (TP); Red: false positives (FP); Black: true negatives (TN); Green: false negatives (FN).

*2) Change maps:* Second, to demonstrate the effectiveness of the proposed VDF-HCD, five recently proposed SOTA methods for HCD are adopted for comparison, including M3CD [55], FPMS [44], CICM [9], NPSG [12] and IRG-McS [36]. We use the default parameters in their codes, which are also consistent with the related papers.

The visual results of the binary CMs generated by different methods on all the heterogeneous data sets are shown in Fig. 5. We can see that some methods do not perform robustly enough and their performance degrades dramatically on some complex HCD scenes, such as M3CD, FPMS, CICM and NPSG on Dataset #2, M3CD on Dataset #6, CICM on Dataset #7. In contrast, the proposed VDF-HCD can achieve robust detection results across different HCD conditions. In the results of Datasets #6 and #7, many false alarms appear in the CM of M3CD; while in the results of Datasets #2 and #3, many missed detection appear in the CM of CICM. On the whole, the proposed VDF-HCD can suppress the false alarms and reduce the missed detection simultaneously, and outperforms other comparison methods. These performances can be attributed to three main factors: 1) the structure difference is robust to different HCD conditions, such as scenes, noises and sensors; 2) the high-order neighborhood information used in the VDF-HCD helps to capture the structure of complex regions accurately and effectively; 3) the negative influences of changes are reduced in the VDF-HCD by cutting off the signal transmission from the changed vertices.

To further illustrate the superiority of our method, the quantitative measures of CMs are reported in Table V. We can see that the proposed method obtains better results than the SOTA methods on most datasets. For example, the VDF-HCD gains the highest OA, Kc and Fm in Datasets #1, #2, #3, #5 and #7. The average OA, Kc and Fm of VDF-HCD on all evaluated data sets are about 0.956, 0.687, and 0.710 respectively, which are higher than other comparison methods. For example, the average Fm of VDF-HCD is 6.6% higher than the second ranked IRG-McS. This demonstrates our VDF based method can efficiently improve the CD performance.

TABLE IV
AUR AND AUP OF DIS GENERATED BY VDF-HCD ON THE HETEROGENEOUS DATA SETS.

| Measures | DIs | Datasets #1 | #2 | #3 | #4 | #5 | #6 | #7 |
|---|---|---|---|---|---|---|---|---|
| AUR | Forward | 0.859 | 0.946 | 0.814 | 0.903 | 0.896 | 0.903 | 0.960 |
| | Backward | 0.900 | 0.937 | 0.805 | 0.951 | 0.971 | 0.897 | 0.923 |
| | Fused | 0.884 | 0.946 | 0.842 | 0.941 | 0.963 | 0.917 | 0.956 |
| AUP | Forward | 0.571 | 0.850 | 0.468 | 0.659 | 0.479 | 0.463 | 0.653 |
| | Backward | 0.615 | 0.800 | 0.521 | 0.768 | 0.810 | 0.284 | 0.609 |
| | Fused | 0.658 | 0.845 | 0.541 | 0.763 | 0.766 | 0.425 | 0.667 |

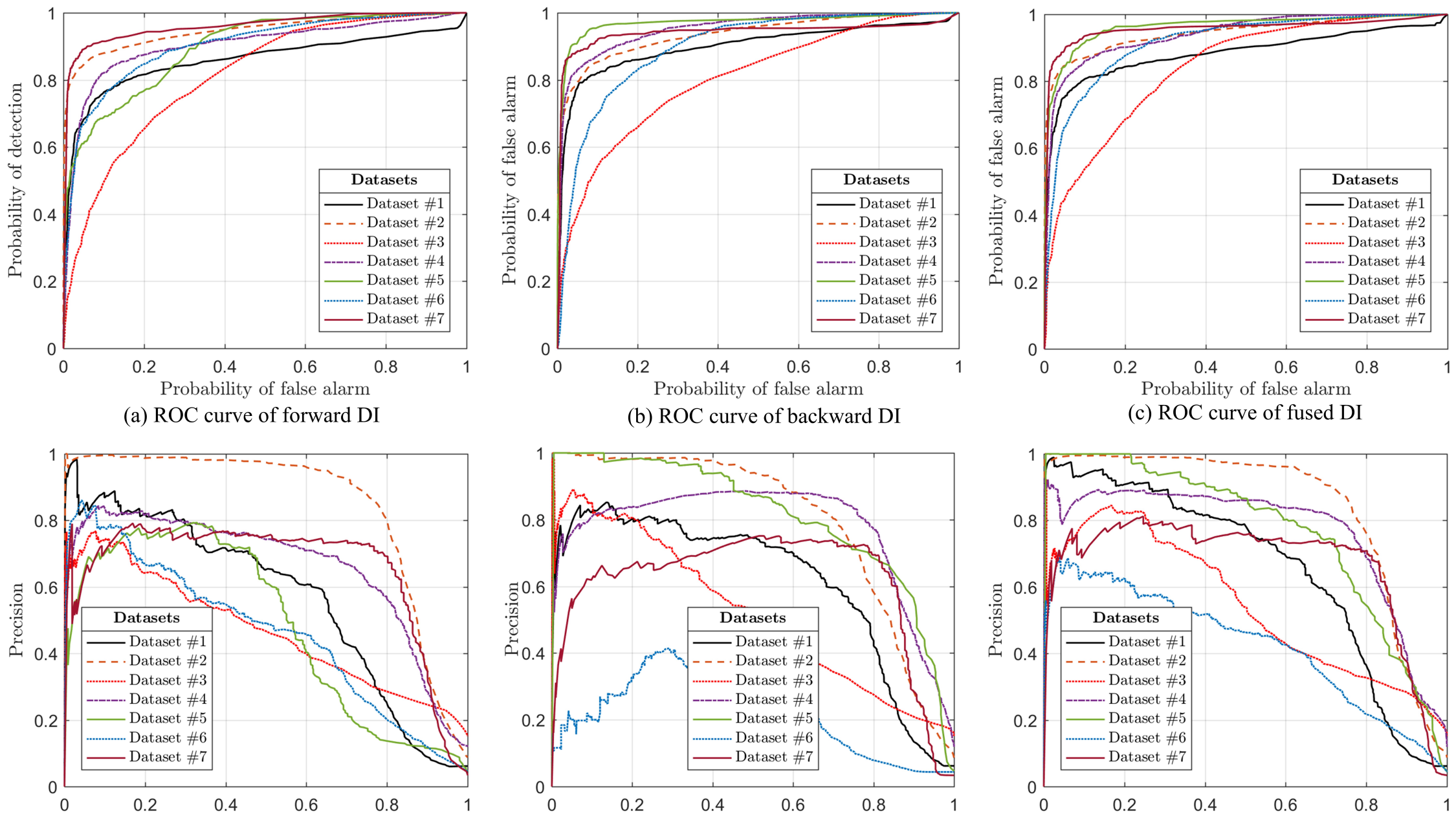

Fig. 6. ROC and PR curves of VDF-HCD generated DIs: (a) ROC curve of forward DI; (b) ROC curve of backward DI; (c) ROC curve of fused DI; (d) PR curve of forward DI; (e) PR curve of backward DI; (f) PR curve of fused DI.

## C. Discussions

*1) The choices of $H(\mathbf{S})$:* We use the polynomials of $H(\mathbf{W}^{\text{avg}})$ to approximate the low-pass filter with the transfer function of $H(\mathbf{\Gamma}) = \frac{sign(\mathbf{\Gamma}-\gamma_{cf})+1}{2}$. It can be proven that the eigenvalues of $\mathbf{W}^{\text{avg}}$ satisfy $\gamma_k \in [-1, 1]$, $k = 1, 2, \cdots, N$, and the larger eigenvalue represents the lower frequency (as illustrated in part II [45]). In Fig. 7, we plot the different polynomial transfer functions to approximate the functions of $H(\mathbf{\Gamma})$ with different cut-off frequency $\gamma_{cf}$. In these filters, two parameters are important: the polynomial order $M$ that corresponds to the hop number of neighbors that each vertex

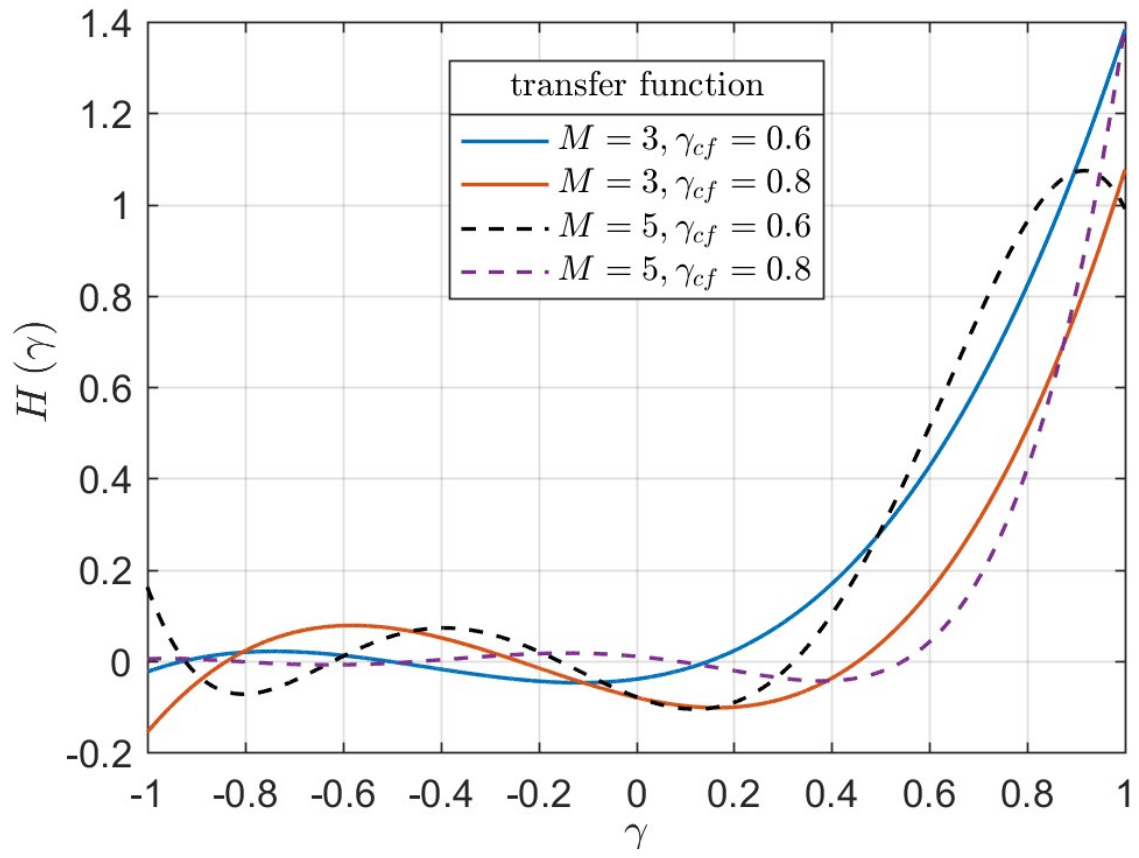

Fig. 7. Polynomial transfer function to approximate the $H(\mathbf{\Gamma})$ with different polynomial orders $M$ and different cut-off frequencies $\gamma_{cf}$.

TABLE V
QUANTITATIVE MEASURES OF BINARY CMS ON THE HETEROGENEOUS DATA SETS. THE HIGHEST SCORES ARE HIGHLIGHTED IN BOLD.

| Datasets | M3CD [55] | | | FPMS [44] | | | CICM [9] | | | NPSG [12] | | | IRG-McS [36] | | | VDF-HCD (proposed) | | |
|---|---|---|---|---|---|---|---|---|---|---|---|---|---|---|---|---|---|---|
| | OA | Kc | Fm | OA | Kc | Fm | OA | Kc | Fm | OA | Kc | Fm | OA | Kc | Fm | OA | Kc | Fm |
| Dataset #1 | 0.963 | 0.669 | 0.689 | 0.925 | 0.552 | 0.588 | 0.943 | 0.451 | 0.481 | 0.947 | 0.559 | 0.587 | **0.971** | 0.739 | 0.754 | **0.971** | **0.756** | **0.771** |
| Dataset #2 | 0.896 | 0.009 | 0.044 | 0.893 | 0.001 | 0.012 | 0.907 | 0.006 | 0.021 | 0.902 | 0.458 | 0.511 | 0.933 | 0.473 | 0.506 | **0.973** | **0.813** | **0.828** |
| Dataset #3 | 0.863 | 0.405 | 0.481 | 0.838 | 0.215 | 0.296 | 0.867 | 0.270 | 0.321 | 0.830 | 0.346 | 0.446 | 0.882 | 0.420 | 0.478 | **0.886** | **0.448** | **0.504** |
| Dataset #4 | 0.915 | 0.588 | 0.636 | **0.962** | **0.816** | **0.837** | 0.884 | 0.507 | 0.573 | 0.902 | 0.608 | 0.663 | 0.939 | 0.714 | 0.749 | 0.944 | 0.754 | 0.786 |
| Dataset #5 | 0.962 | 0.602 | 0.622 | 0.938 | 0.569 | 0.597 | 0.974 | 0.745 | 0.759 | 0.975 | 0.729 | 0.742 | 0.983 | 0.794 | 0.804 | **0.985** | **0.808** | **0.816** |
| Dataset #6 | 0.575 | 0.021 | 0.077 | 0.947 | 0.329 | 0.356 | 0.899 | 0.081 | 0.131 | 0.941 | 0.419 | 0.449 | **0.959** | **0.490** | **0.512** | 0.952 | 0.487 | **0.512** |
| Dataset #7 | 0.856 | 0.158 | 0.204 | 0.979 | 0.544 | 0.553 | 0.789 | 0.024 | 0.080 | 0.985 | 0.733 | 0.741 | 0.976 | 0.690 | 0.702 | **0.982** | **0.745** | **0.754** |
| Average | 0.861 | 0.350 | 0.393 | 0.926 | 0.432 | 0.463 | 0.895 | 0.300 | 0.338 | 0.926 | 0.550 | 0.591 | 0.949 | 0.617 | 0.644 | **0.956** | **0.687** | **0.710** |

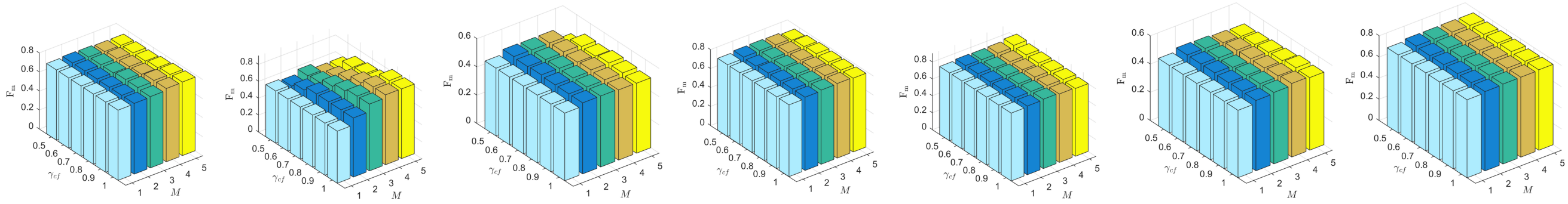

(a) Dataset #1 (b) Dataset #2 (c) Dataset #3 (d) Dataset #4 (e) Dataset #5 (f) Dataset #6 (g) Dataset #7

Fig. 8. Sensitivity analysis of $H(\mathbf{\Gamma})$ with different $M$ and $\gamma_{cf}$ in VDF-HCD on all heterogeneous data sets.

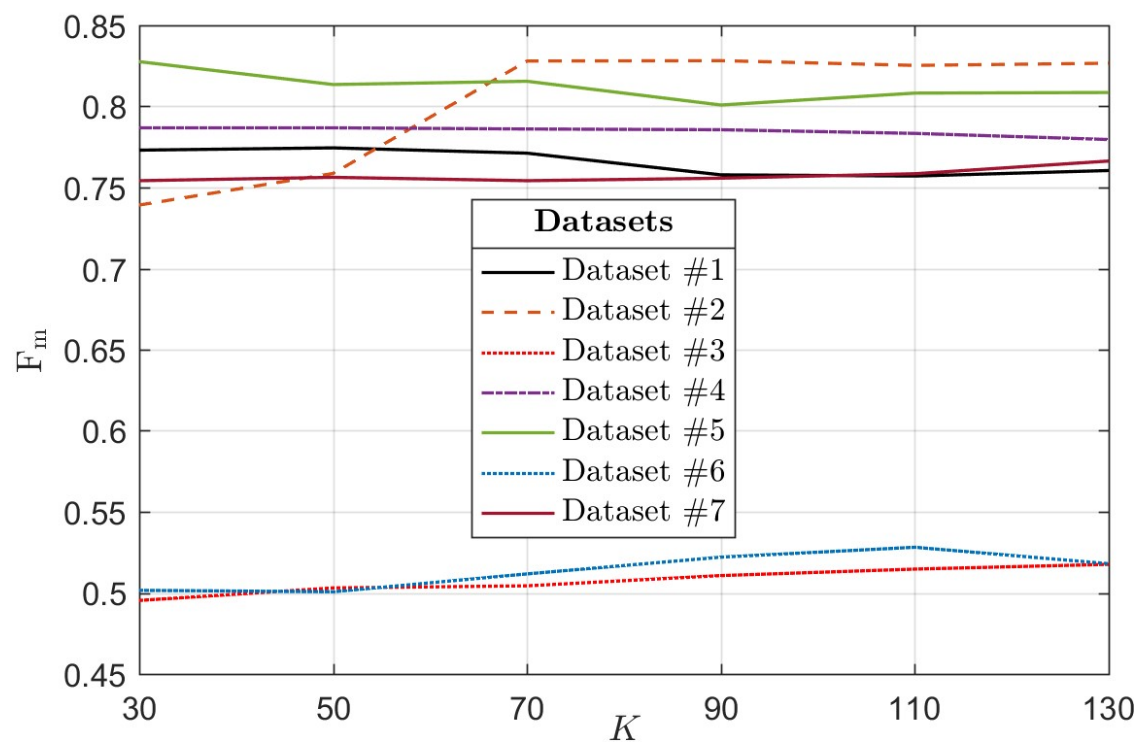

Fig. 9. Sensitivity analysis of parameter $K$ in VDF-HCD.

can reach, the cut-off frequency that determines the bandwidth of the filter.

In Fig. 8, we verify the polynomial order $M$ (from 1 to 5) and cut-off frequency $\gamma_{cf}$ (from 0.5 to 1 with an interval of 0.1) in the proposed VDF-HCD. Two remarks can be observed in Fig. 8: 1) the higher order $M$ usually brings better results than the $M = 1$ (the average Fm value is only 0.658 when $M = 1$), which demonstrates the effectiveness of the high-order information. However, a larger $M$ is also not recommend for two reasons: first, creating redundancy; second, computational complex. 2) The $\gamma_{cf}$ controls the bandwidth of the low-pass filter, as shown in Fig. 7. Although it can be estimated accurately by the graph Fourier transform as illustrated by Fig. 2 of part II [45], this process requires eigenvalue decomposition, which is very complex for large scale graph. However, as can be seen from the Fig. 8 that the method is robust for parameter $\gamma_{cf}$, there is a wide range of acceptable values for $\gamma_{cf}$. For example, in our experiments we fixed $\gamma_{cf} = 0.9$ for simplicity.

*2) The choices of K:* In the KNN graph, the number of nearest neighbors plays an important role. In Fig. 9, we vary $K$ from 30 to 130 with step 20. It can be found that the detection performance is not very sensitive to the value of $K$. Of cause, a very small $K$ is not appropriate, which will make the graph not robust and thus affect the change measurements. For example, for the unchanged vertex $i$, when one of its neighbor $j' \in \mathcal{N}_i^{\mathbf{y}}$ is polluted by noise or changed, the $\mathbf{X}_{j'}$ will be different from the $\mathbf{X}_i$ and it brings errors in the $\mathbf{d}_i^{\mathbf{x}}$ of (8). In this case, a larger $K$ will reduce the influence of $\mathbf{X}_{j'}$ by average weighting. This can also be illustrated by Fig. 1(a) and Fig. 3(a), where the unchanged 4th and 7th vertices with three neighbors in $G_{t1}$ and $G_{t2}$ are less affected by changed 5th vertex than the others, *e.g.*, the 1st, 2nd vertices in $G_{t1}$, and the 6th, 9th vertices in $G_{t2}$, with two neighbors. On the other hand, a particularly large $K$ is also not appropriate, which will make the change measurements less discriminative. For example, when the $K$ is exceeds the actual number of the real similar neighbors of the vertex, then this vertex will receive information from the vertices that are not related to it. In the extreme case of $K = N$, we have the measurement of (8) equals to 0 for all vertices. Therefore, we empirically set $K = \sqrt{N}$, which comes from the work of KNN based density estimation [56]–[58] and KNN classification [59].

*3) The effect of reducing influence of changes:* In order to avoid propagating the influence of changed vertex to their neighboring vertices, we use an iterative framework to eliminate the changed vertices detected by the previous round in the graph construction of the next round, as illustrated by Table II and Fig. 4(d). To verify the effectiveness of this propagation cut-off, we further investigate the process of DI calculation. Figure 10(a) and 10(b) show the proportions of changed vertices in the KNN of each vertex in the graphs of $G_{t1}$ and $G_{t2}$, respectively. As can be seen, some vertices in $G_{t2}$ contains a lot of changed vertices in their NNs, as shown in the middle of Fig. 10(b). These changed vertices in turn affect the calculation of $\mathbf{f}^{\mathbf{x}}$ (14), making it less discriminative. Therefore, the initial forward and backward DIs without the change elimination perform very differently, as shown in Fig. 10(c) and 10(d). Once we remove the influence of changed vertices, we can find that DIs perform much better, as illustrated by the forward and backward DIs generated by the final iteration of VDF-HCD in Fig. 10(e) and 10(f). From Fig. 10(c)-(f), it is clear that cutting off signal propagation from changed vertices is very effective.

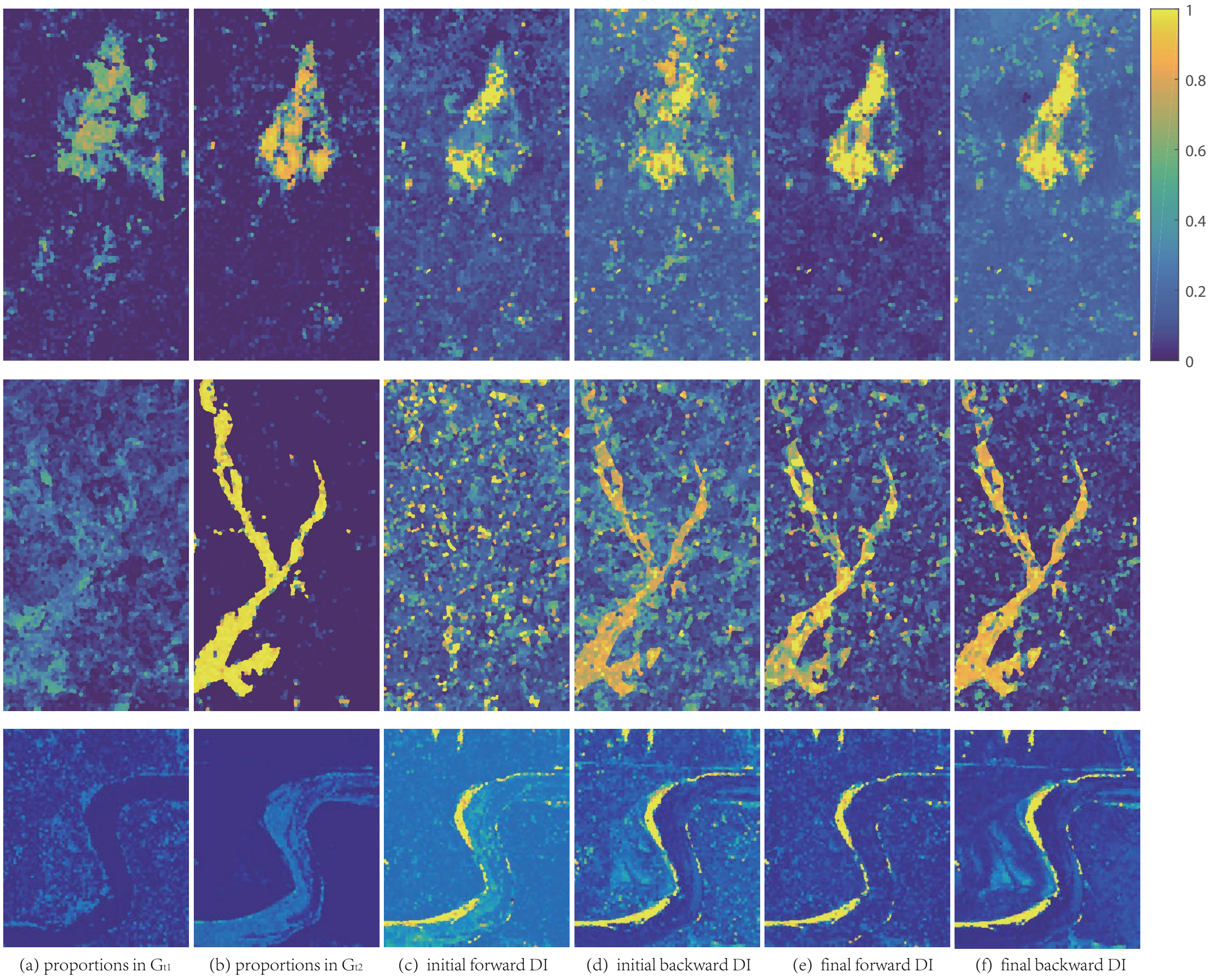

Fig. 10. DIs generated by the initial and final iterations of VDF-HCD on Datasets #2, #4 and #7. From top to bottom, they correspond to Datasets #2, #4 and #7, respectively. From left to right are: (a)-(b) the proportions of changed vertices in the KNN of each vertex in the graphs of $G_{t1}$ and $G_{t2}$, respectively; (c)-(d) the forward and backward DIs generated by the initial iteration of VDF-HCD, respectively; (e)-(f) the forward and backward DIs generated by the final iteration of VDF-HCD, respectively.

*4) Computational analysis:* The main computational complexity of the proposed method is concentrating on the DI generation and MRF co-segmentation. For the former, calculating the distance matrix between all the patches or superpixels requires $\mathcal{O}\left(\left(M_x+M_y\right)N^2/2\right)$, sorting the distance matrix by column to construct the KNN graph requires $\mathcal{O}\left(N^2 \log N\right)$, calculating the $H\left(\mathbf{S}\right)$ requires $\mathcal{O}\left(\left(M-1\right)N^3\right)$. For the MRF co-segmentation, it requires $\mathcal{O}\left(2N_R N^2\right)$ with the worst-case [36], where $N_R$ is the number of edges in the $R$-adjacency neighbor system of the MRF co-segmentation model.

Table VI reports the computational time of each process of VDF-HCD with different $N$ on Datasets #1 and #3, which is performed in MATLAB 2016a running on a Windows desktop with Intel Core i7-8700K CPU. In Table VI, $t_{pre}$, $t_{di}$, and $t_{seg}$ represent the computational times spent in the pre-processing, DI calculation, and MRF co-segmentation respectively, $t_{total}$ represents the total running time. As can be seen in Table VI, the running time of VDF-HCD is mainly determined by the graph scale $N$, and DI calculation is the most time-consuming process in VDF-HCD.

TABLE VI
COMPUTATIONAL TIME (SECONDS) OF EACH PROCESS OF VDF-HCD.

| Data sets | $N$ | $t_{pre}$ | $t_{di}$ | $t_{seg}$ | $t_{total}$ |
|---|---|---|---|---|---|
| Dataset #7<br>$343\times 291\times 1(3)$ | 5000<br>10000 | 0.44<br>1.08 | 19.96<br>82.69 | 5.39<br>15.41 | 26.49<br>100.13 |
| Dataset #2<br>$2000\times 2000\times 3(3)$ | 5000<br>10000 | 3.15<br>3.77 | 20.50<br>83.09 | 5.87<br>15.92 | 29.91<br>104.33 |

## VI. CONCLUSION

Motivated by the performance gained by the graph based methods, we investigate the inner workings of graph based HCD methods, and propose a new strategy for solving the HCD problem from the perspective of graph signal processing. By defining the graph and graph signal, the changes between heterogeneous images manifest themselves in two aspects: the structure difference between graphs and the signal difference on the graph. Thereby, we can compare the responses of the two signals on different filters to detect the changes.

In this part, we propose a framework to solve the HCD based on the GSP with vertex domain filtering, which measures the structure difference between graphs by comparing the output signals of the same input signal on filters defined on different graphs. The proposed VDF-HCD can explore the high-order information hidden in the graphs by using different filters, and alleviate the negative influence of changes by using an iterative strategy that cuts off the signal propagation from changed vertices to their neighboring vertices. Experimental results on seven heterogeneous data sets demonstrate the effectiveness of VDF-HCD. In the proposed method, the graph filter plays an important role. A future work is to design better filters to improve detection accuracy.

## Acknowledgment

The authors would like to thank the researchers for their friendly sharing of heterogeneous change detection codes and data sets, which provide a wealth of resources for this study.